\documentclass{article}

\usepackage[numbers, compress]{natbib}  
\usepackage[nonatbib,final]{neurips_2022}
\usepackage{tikz}
\usepackage{comment}
\usepackage{amsmath,amssymb} 
\usepackage{color}
\usepackage{amsmath}
\usepackage{amssymb}
\usepackage{mathtools}
\usepackage{wrapfig}

\usepackage{graphicx}
\usepackage{amsmath,amsfonts,bm}
\usepackage{booktabs}
\usepackage{algorithm}
\usepackage{algorithmic}

\usepackage{bbding}
\usepackage{multirow}
\usepackage{subfigure}
\usepackage{booktabs}
\usepackage{array}
\usepackage[flushleft]{threeparttable}
\usepackage{xspace}

\usepackage{amsmath,amsfonts,bm}

\def\eqref#1{equation~\ref{#1}}

\def\1{\bm{1}}

\def\rvh{{\mathbf{h}}}

\def\rvy{{\mathbf{y}}}

\def\rmE{{\mathbf{E}}}

\DeclareMathAlphabet{\mathsfit}{\encodingdefault}{\sfdefault}{m}{sl}
\SetMathAlphabet{\mathsfit}{bold}{\encodingdefault}{\sfdefault}{bx}{n}

\def\sR{{\mathbb{R}}}

\def\bs{\bm}

\DeclarePairedDelimiterX{\infdivx}[2]{(}{)}{%
  #1\;\delimsize|\delimsize|\;#2%
}
\newcommand{\kld}[2]{\ensuremath{D_{KL}\infdivx{#1}{#2}}\xspace}

\usepackage[utf8]{inputenc} 
\usepackage[T1]{fontenc}    
\usepackage{hyperref}       
\usepackage{url}            
\usepackage{booktabs}       
\usepackage{amsfonts}       
\usepackage{nicefrac}       
\usepackage{microtype}      
\usepackage{xcolor}         

\title{Your ViT is Secretly a Hybrid Discriminative-Generative Diffusion Model}

\author{%
  Xiulong Yang\thanks{This work was done during an internship at Etsy} \\
  Georgia State University\\
  \texttt{xyang22@gsu.edu} \\
  \And
  Sheng-Min Shih, Yinlin Fu, Xiaoting Zhao \\
  Etsy Inc. \\
  \texttt{\{sshih,yfu,xzhao\}@etsy.com} \\
  \AND
  Shihao Ji \\
  Georgia State University \\
  \texttt{sji@gsu.edu} \\
}

\begin{document}

\maketitle

\begin{abstract}
Diffusion Denoising Probability Models (DDPM)~\cite{DDPM} and Vision Transformer (ViT)~\cite{ViT2021} have demonstrated significant progress in generative tasks and discriminative tasks, respectively, and thus far these models have largely been developed in their own domains. In this paper, we establish a direct connection between DDPM and ViT by integrating the ViT architecture into DDPM, and introduce a new generative model called Generative ViT (GenViT). The modeling flexibility of ViT enables us to further extend GenViT to hybrid discriminative-generative modeling, and introduce a Hybrid ViT (HybViT). Our work is among the first to explore a single ViT for image generation and classification jointly. We conduct a series of experiments to analyze the performance of proposed models and demonstrate their superiority over prior state-of-the-arts in both generative and discriminative tasks. Our code and pre-trained models can be found in \url{https://github.com/sndnyang/Diffusion_ViT}.
\end{abstract}

\section{Introduction}

Discriminative models and generative models based on the Convolutional Neural Network (CNN)~\cite{CNN} architectures, such as GAN~\cite{GAN} and ResNet~\cite{resnet16}, have achieved state-of-the-art performance in a wide range of learning tasks. Thus far, they
have largely been developed in two separate domains. In recent years, ViTs have started to rival CNNs in many vision tasks. Unlike CNNs, ViTs can capture the features from an entire image by self-attention, and they have demonstrated superiority in modeling non-local contextual dependencies as well as their efficiency and scalability to achieve comparable classification accuracy with smaller computational budgets (measured in FLOPs). Since the inception, ViTs have been exploited in various tasks such as object detection~\cite{carion2020detr}, video recognition~\cite{bertasius2021space}, multi-modal pre-training~\cite{kim2021vilt}, and image generation~\cite{jiang2021transgan,vitgan2022}. Especially, VQ-GAN~\cite{esser2021vqgan}, TransGAN~\cite{jiang2021transgan} and ViTGAN~\cite{vitgan2022} investigate the application of ViT in image generation. However, VQ-GAN is built upon an extra CNN-based VQ-VAE, and the latter two require two ViTs to construct a GAN for generation tasks. Therefore we ask the following question: is it possible to train a generative model using a single ViT? 

DDPM is a class of generative models that matches a data distribution by learning to reverse a multi-step diffusion process. It has recently been shown that DDPMs can even outperform prior SOTA GAN-based generative models~\cite{ddpmbeatgan2021,biggan,styleganADA}. Unlike GAN which needs to train with two competing networks, DDPM utilizes a UNet~\cite{ronneberger2015unet} as a backbone for image generation and is trained to optimize maximum likelihood to avoid the notorious instability issue in GAN~\cite{miyato2018spectral,biggan} and EBM~\cite{du2019implicit,jem}. 

In this paper, we establish a direct connection between DDPM and ViT for the task of image generation and classification. Specifically, we answer the question whether a single ViT can be trained as a generative model. We design \textbf{Gen}erative ViT (GenViT) for pure generation tasks, as well as \textbf{Hyb}rid ViT (HybViT) that extends GenViT to a hybrid model for both image classification and generation. As shown in Fig~\ref{fig:arch_genvit} and~\ref{fig:arch_hybvit}, the reconstruction of image patches and the classification are two routines independent to each other and train a shared set of features together.

Our experiments show that HybViT outperforms previous state-of-the-art hybrid models. In particular, the Joint Energy-based Model (JEM), the previous state-of-the-art proposed by \cite{jem,jempp}, requires extremely expensive MCMC sampling, which introduce instability and causes the training processes to fail for large-scale datasets due to the long training procedures required. To the best of our knowledge, GenViT is the first model that utilizes a single ViT as a generative model, and HybViT is a new type of hybrid model without the expensive MCMC sampling during training.
Compared to existing methods, our new models demonstrate a number of conceptual advantages~\cite{du2019implicit}: 1) Our methods provide simplicity and stability similar to DDPM, and are less prone to collapse compared to GANs and EBMs. 2) The generative and discriminative paths of our model are trained with a single objective which enables sharing of statistical strengths. 3) Advantageous computational efficiency and scalability to growing model and data sizes inherited from the ViT backbone.

Our contributions can be summarized as following: 

\begin{enumerate}

\item We propose GenViT, which to the best of our knowledge, is the first approach to utilize a single ViT as an alternative to the UNet in DDPM. 
\item We introduce HybViT, a new hybrid approach for image classification and generation leveraging ViT, and show that HybViT considerably outperforms the previous state-of-the-art hybrid models on both classification and generation tasks while at the same time optimizes more effectively than MCMC-based models such as JEM/JEM++.
\item We perform comprehensive analysis on model characteristics including adversarial robustness, uncertainty calibration, likelihood and OOD detection, comparing GenViT and HybViT with existing benchmarks.

\end{enumerate}

\section{Related Work}

\subsection{Denoising Diffusion Probabilistic Models}

\label{sec:ddpm}
\begin{figure}[h]
    \centering
        \includegraphics[width=0.98\columnwidth]{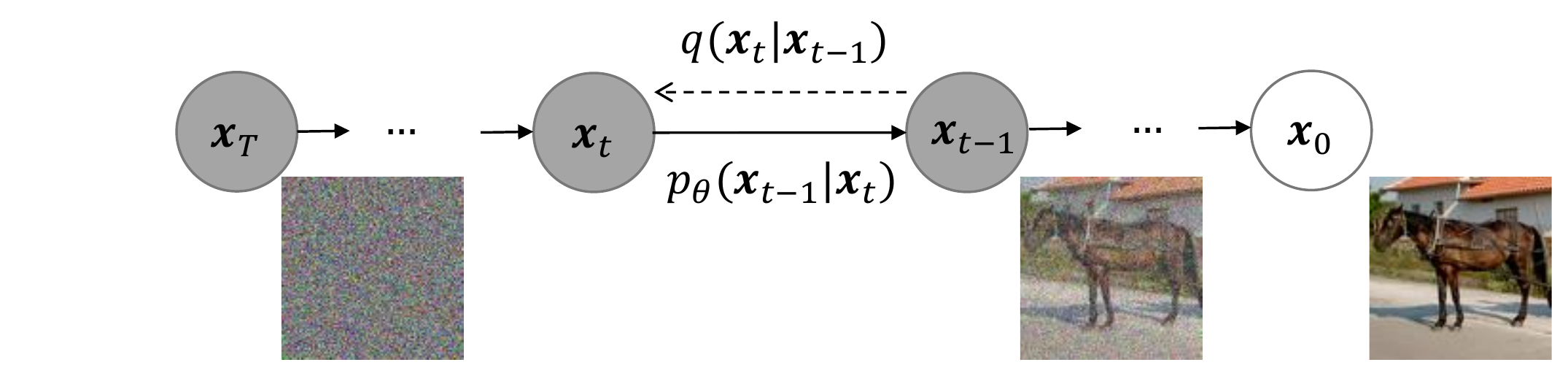}
    \caption{A graphical model of diffusion process.}
    \label{fig:ddpm}
\end{figure}

We first review the derivation of DDPM~\cite{DDPM}. DDPM is built upon the theory of Nonequilibrium Thermodynamics~\cite{sohl-dickstein2015deep} with a few simple yet effective assumptions. It assumes diffusion is a noising process $q$ that accumulates isotropic Gaussian noises over timesteps (Figure~\ref{fig:ddpm}).

Starting from the data distribution $\vec{x}_0 \sim q(\vec{x}_0)$, the diffusion process $q$ produces a sequence of latents $\vec{x}_1$ through $\vec{x}_T$ by adding Gaussian noise at each time $t \in [0, \cdots, T-1]$ with variance $\beta_t \in (0,1)$ as follows:
\begin{alignat}{2}
    q(\vec{x}_1, ..., \vec{x}_T | \vec{x}_0) &\coloneqq \prod_{t=1}^{T} q(\vec{x}_t | \vec{x}_{t-1}) \label{eq:joint} \\
    q(\vec{x}_t | \vec{x}_{t-1}) &\coloneqq \mathcal{N}(\vec{x}_t; \sqrt{1-\beta_t} \vec{x}_{t-1}, \beta_t \mathbf{I}) \label{eq:singlestep}
\end{alignat}

Then, the process in reverse aims to get a sample in $q(\vec{x}_0)$ from sampling $\vec{x}_T \sim \mathcal{N}(0, \mathbf{I})$ by using a neural network:
\begin{equation}
\label{eq:nn}
p_{\theta}(\vec{x}_{t-1}|\vec{x}_t) \coloneqq \mathcal{N}(\vec{x}_{t-1}; \mu_{\theta}(\vec{x}_t, t), \Sigma_{\theta}(\vec{x}_t, t))
\end{equation}

With the approximation of $q$ and $p$, DDPM gets a variational lower bound (VLB) as follows:
\begin{align}\label{eq:elbo}
    \log p_{\bs{\theta}(\bs{x}_0)} & \geq  \log p_{\bs{\theta}(\bs{x}_0)} - \kld{ q(\bs{x}_{1:T} | \bs{x}_0)}{p_{\bs{\theta}} (\bs{x}_{0:T}) } \nonumber \\
    & = - \mathbb{E}_q \left[  \frac{q(\bs{x}_{1:T} | \bs{x}_0)}{p_{\bs{\theta}} (\bs{x}_{0:T})} \right] 
\end{align}

Then they derive a loss for VLB as:
\begin{alignat}{2}
    L_{\text{vlb}} &= L_0 + L_1 + ... + L_{T-1} + L_T \label{eq:loss} \\
    L_{0} &= -\log p_{\theta}(\vec{x}_0 | \vec{x}_1) \label{eq:loss0} \\
    L_{t-1} &= \kld{q(\vec{x}_{t-1}|\vec{x}_t,\vec{x}_0)}{p_{\theta}(\vec{x}_{t-1}|\vec{x}_t)} \label{eq:losst} \\
    L_{T} &= \kld{q(\vec{x}_T | \vec{x}_0)}{p(\vec{x}_T)} \label{eq:lossT} 
\end{alignat}
where $L_0$ is modeled by an independent discrete decoder from the Gaussian $\mathcal{N}(\vec{x}_{0}; \mu_{\theta}(\vec{x}_1, 1), \sigma_{1}^{2} \vec{I})$, and $L_T$ is constant and can be ignored.

As noted in \cite{DDPM}, the forward process can sample an arbitrary timestep $\bs{x}_t$ directly conditioned on the input $\bs{x}_0$ in a closed form. With the nice property, we define $\alpha_t \coloneqq 1 - \beta_t$ and $\bar{\alpha}_t \coloneqq \prod_{s=0}^{t} \alpha_s$. Then we have 
\begin{alignat}{2}
    q(\vec{x}_t|\vec{x}_0) &= \mathcal{N}(\vec{x}_t; \sqrt{\bar{\alpha}_t} \vec{x}_0, (1-\bar{\alpha}_t) \mathbf{I}) \label{eq:marginal} \\
    \vec{x}_t &=  \sqrt{\bar{\alpha}_t} \vec{x}_0 + \sqrt{1-\bar{\alpha}_t} \epsilon \label{eq:0_to_t}
\end{alignat}
where $\epsilon\!\!\sim\!\!\mathcal{N}(0,\mathbf{I})$ using the reparameterization.  Then using Bayes theorem, we can calculate the posterior $q(\vec{x}_{t-1}|\vec{x}_t,\vec{x}_0)$ in terms of $\tilde{\beta}_t$ and $\tilde{\mu}_t(\vec{x}_t,\vec{x}_0)$ as follows:
\begin{align}
    q(\vec{x}_{t-1}|\vec{x}_t,\vec{x}_0) &= \mathcal{N}(\vec{x}_{t-1}; \tilde{\mu}(\vec{x}_t, \vec{x}_0), \tilde{\beta}_t \mathbf{I}) \label{eq:posterior} \\
    \tilde{\mu}_t(\vec{x}_t,\vec{x}_0) &\coloneqq\! \frac{\sqrt{\bar{\alpha}_{t-1}}\beta_t}{1-\bar{\alpha}_t}\vec{x}_0\!+\! \frac{\sqrt{\alpha_t}(1\!-\!\bar{\alpha}_{t-1})}{1\!-\!\bar{\alpha}_t} \vec{x}_t \label{eq:mu_tilde} \\
    \tilde{\beta}_t &\coloneqq \frac{1-\bar{\alpha}_{t-1}}{1-\bar{\alpha}_t} \beta_t \label{eq:beta_tilde}
\end{align}

As we can observe, the objective in Eq.~\ref{eq:loss} is a sum of independent terms $L_{t-1}$. Using Eq.~\ref{eq:0_to_t}, we can sample from an arbitrary step of the forward diffusion process and estimate $L_{t-1}$ efficiently. Hence, DDPM uniformly samples $t$ for each sample in each mini-batch to approximate the expectation $E_{\bs{x}_0,t,\epsilon}[L_{t-1}]$ to estimate $L_{\text{vlb}}$.

To parameterize $\mu_{\theta}(\vec{x}_t, t)$ for Eq.~\ref{eq:mu_tilde}, we can predict $\mu_{\theta}(\vec{x}_t, t)$ directly with a neural network. Alternatively, we can first use Eq.~\ref{eq:0_to_t} to replace $\bs{x}_0$ in Eq.~\ref{eq:mu_tilde} to predict the noise $\epsilon$ as
\begin{equation}
\mu_{\theta}(\vec{x}_t, t) = \frac{1}{\sqrt{\alpha_t}} \left( \vec{x}_t - \frac{\beta_t}{\sqrt{1-\bar{\alpha}_t}} \epsilon_{\theta}(\vec{x}_t, t) \right), \label{mu_to_noise}
\end{equation}
\cite{DDPM} finds that predicting the noise  $\epsilon$ worked best with a reweighted loss function:
\begin{equation}\label{eq:loss_simple}
L_{\text{simple}} = E_{t,\vec{x}_0,\epsilon}\left[ || \epsilon - \epsilon_{\theta}(\vec{x}_t, t) ||^2 \right].
\end{equation}
This objective can be seen as a reweighted form of $L_{\text{vlb}}$ (without the terms affecting $\Sigma_{\theta}$). For more details of the training and inference, we refer the readers to~\cite{DDPM}. A closely related branch is called score matching~\cite{song2019generative,song2020score}, which builds a connection bridging DDPMs and EBMs. Our work is mainly built upon DDPM, but it's straightforward to substitute DDPM with a score matching method.

\section{Vision Transformers}
\label{transformer}

Transformers \cite{vaswani2017attention} have made huge impacts across many deep learning fields \cite{han2020survey} due to their prediction power and flexibility. They are based on the concept of self-attention, a function that allows interactions with strong gradients between all inputs, irrespective of their spatial relationships. The self-attention layer (Eq.~\ref{eqn:attention}) encodes inputs as key-value pairs, where values $\vec{V}$ represent embedded inputs and keys $\vec{K}$ act as an indexing method, and subsequently, a set of queries $\vec{Q}$ are used to select which values to observe. Hence, a single self-attention head is computed as:
\begin{equation}\label{eqn:attention}
    \text{Attn}(\vec{Q}, \vec{K}, \vec{V}) = \text{softmax}\bigg( \frac{\vec{Q}\vec{K}^T}{\sqrt{d_k}} \bigg) \vec{V}.
\end{equation}
where $d_k$ is the dimension of $K$.

\begin{figure*}[ht]
    \centering
        \includegraphics[width=0.98\textwidth]{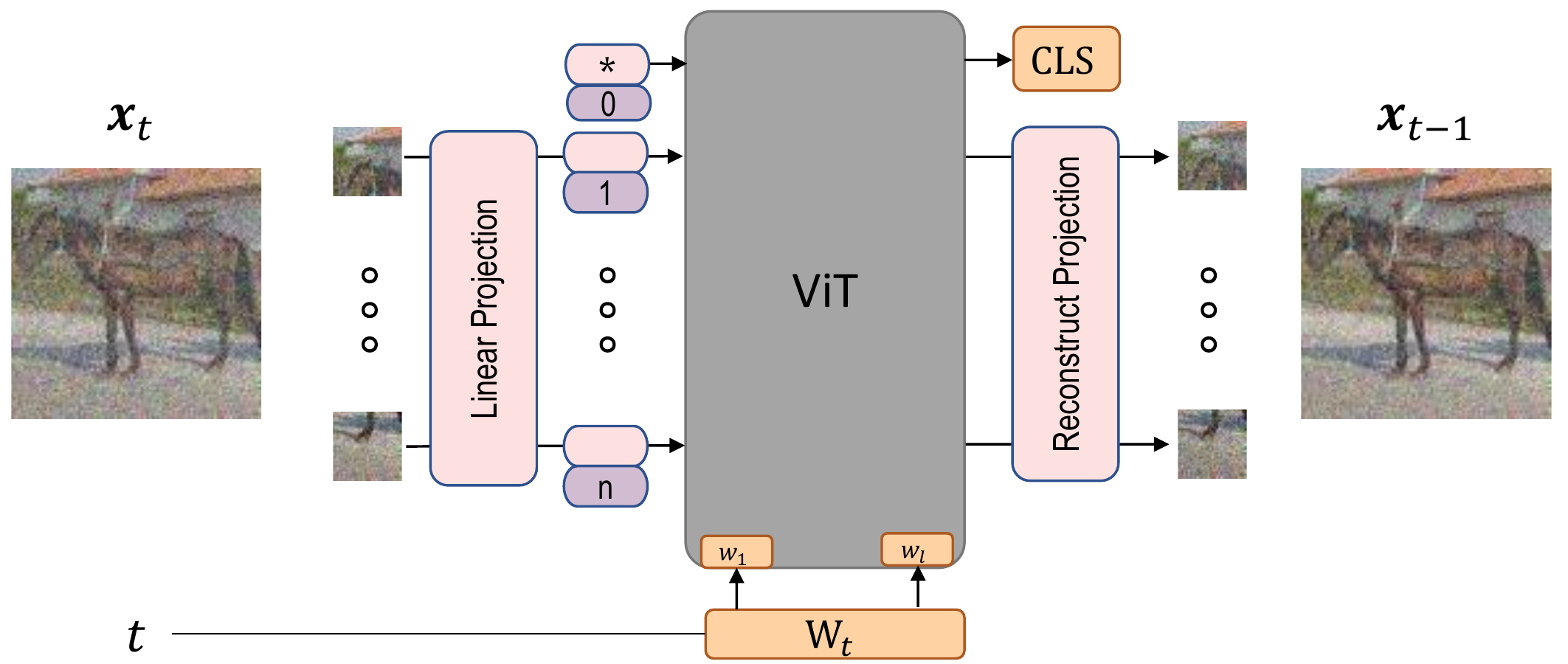}
    \caption{The backbone architecture for GenViT and HybViT. For generative modeling, $\bs{x}_t$ with a time embedding of $t$ is fed into the model. For the classification task in HybViT, we compute logits from CLS with the input $\bs{x}_0$.}
    \label{fig:arch_genvit}
\end{figure*}

Vision transformers (ViT) {ViT2021} has emerged as a famous architecture that outperforms CNNs in various vision domains. The transformer encoder is constructed by alternating layers of multi-headed self-attention (MSA) and MLP blocks (Eq.~\ref{eq:msa_apply}, \ref{eq:mlp_apply}), and layernorm (LN) is applied before every block, followed by residual connections after every block \cite{wang2019-preLN,Baevski2019Adaptive}.
The MLP contains two layers with a GELU non-linearity. The 2D image  $\bs{x} \in \sR^{H \times W \times C}$ is flattened into a sequence of image patches, denoted by $\bs{x}_p \in \sR^{L \times (P^2 \cdot C)}$, where $L=\frac{H\times W}{P^2}$ is the effective sequence length and $P \times P \times C$ is the dimension of each image patch. 

Following BERT~\cite{bert18}, we prepend a learnable classification embedding $\bs{x}_\text{class}$ to the image patch sequence, then the 1D positional embeddings $\rmE_{pos}$ are added to formulate the patch embedding $\bs{z}_0$. The overall pipeline of ViT is shown as follows:
\begin{align}
    \bs{z}_0 = &[ \bs{x}_\text{class}; \, \bs{x}^1_p \bs{E}; \, \bs{x}^2_p \bs{E}; \cdots; \, \bs{x}^{N}_p \bs{E} ] + \bs{E}_{pos},  \label{eq:embedding} \\
    & \bs{E} \in \mathbb{R}^{(P^2 \cdot C) \times D},\, \bs{E}_{pos}  \in \mathbb{R}^{(N + 1) \times D}  \nonumber \\
    \bs{z^\prime}_\ell = &\text{MSA}(\text{LN}(\bs{z}_{\ell-1})) + \bs{z}_{\ell-1},  \;\; \ell=1\ldots L \label{eq:msa_apply} \\
    \bs{z}_\ell = &\text{MLP}(\text{LN}(\bs{z^\prime}_{\ell})) + \bs{z^\prime}_{\ell}, \;\; \;\; \;\;  \ell=1\ldots L  \label{eq:mlp_apply} \\
    \bs{y} = &\text{LN}(\bs{z}_L^0) \label{eq:final_rep}
\end{align}

ViT have made significant breakthroughs in various discriminative tasks and generative tasks, including image classification, multi-modal, and high-quality image and text generation~\cite{ViT2021,METER2022,vitgan2022}. Inspired by the  parallelism between patches/embeddings of ViT, we experiment with applying a standard ViT directly to generative modeling with minimal possible modifications. 

\subsection{Hybrid models}
Hybrid models~\cite{raina2004classification} commonly model the  density function $p(\bs{x})$ and perform discriminative classification jointly using shared features.
Notable examples are~\cite{sslgenerative14,chongxuan2017triple,nalisnick2019hybrid,jem,nomcmc,generalebm}.

Hybrid models can utilize two or more classes of generative model to balance the trade-off such as slow sampling and poor scalability with dimension. For example, VAE can be increased by applying a second generative model such as a Normalizing Flow~\cite{kingma2016improved,grathwohl2018ffjord,vahdat2020nvae} or EBM~\cite{pang2020learning} in latent space. Alternatively, a second model can be used to correct samples~\cite{vaebm}. In our work, we focus on training a single ViT as a hybrid model without the auxiliary model.

\subsection{Energy-Based Models}

Energy-based models (EBMs) are an appealing family of models to represent data as they permit unconstrained architectures. Implicit EBMs define an unnormalized distribution over data typically learned through contrastive divergence~\cite{du2019implicit,hinton2002cd}.

Joint Energy-based Model (JEM)~\cite{jem} reinterprets the standard softmax classifier as an EBM and trains a single network to achieve impressive hybrid discriminative-generative performance. Beyond that, JEM++~\cite{jempp} proposes several training techniques to improve JEM's accuracy, training stability, and speed, including proximal gradient clipping, YOPO-based SGLD sampling, and informative initialization. Unfortunately, training EBMs using SGLD sampling is still impractical for high-dimensional data.

\section{Method}

\label{sec:method}
\subsection{A Single ViT is a Generative Model}

We propose GenViT by substituting UNet, the backbone of DDPM, with a single ViT. In our model design, we follow the standard ViT~\cite{ViT2021} as close as possible. An overview of the architecture of the proposed GenViT is depicted in Fig~\ref{fig:arch_genvit}. 

Given the input $\bs{x}_t$ from DDPM, we follow the raster scan to get a sequence of image patches $\bs{x}_p$, which is fed into GenViT as: 
\begin{align}
    \rvh_0 &= [ \bs{x}_\text{class}; \, \bs{x}_p^1 \rmE; \, \bs{x}_p^2 \rmE; \cdots; \, \bs{x}_p^N \rmE ] + \rmE_{pos},     \nonumber    \\
    & \rmE \in \sR^{(P^2 \cdot C) \times D},\, \rmE_{pos}  \in \sR^{(N + 1) \times D} \nonumber  \\
    \mathbf{h^\prime}_\ell &= \text{MSA}(\text{LN}(M(\mathbf{h}_{\ell-1}, \mathbf{A} ))) + \mathbf{h}_{\ell-1},  \ell=1,\ldots,L    \nonumber \\
    \mathbf{h}_\ell &= \text{MLP}(\text{LN}(M(\mathbf{h^\prime_{\ell}, A}))) + \mathbf{h^\prime}_{\ell},  \ell=1,\ldots,L  \nonumber \\
    \rvy &= \rvh_L = [\rvy^1, \cdots, \rvy^N]   \nonumber \\
    \bs{x}' &= [\bs{x}_p^1, \cdots, \bs{x}_p^N]  = [f_r(\rvy^1),\ldots,f_r(\rvy^N)],  \label{eq:generator_final_rep}  \\ 
     & \bs{x}_p^i \in \sR^{P^2 \times C},  \bs{x}' \in \sR^{H \times W \times C}. \nonumber
\end{align}
Different from ViT, GenViT takes the embedding of $t$ as input to control the hidden features $h_{\ell}$ every layer, and finally reconstruct $L$-th layer output $\bs{h}_L  \in \sR^{(N + 1) \times D}$ to an image $\bs{x}'$. Following the design of UNet in DDPM, we first compute the embedding of $t$ using an MLP $\bs{A} = \text{MLP}_t(t)$. Then we compute $M( \mathbf{h_{\ell}, A} ) = \mathbf{h}_{\ell} * (\mu_{\ell}(\mathbf{A}) + 1) + \sigma_{\ell}(\mathbf{A})$ for each layer, where ${\mu_{\ell}}(\mathbf{A}) = \text{MLP}_{\ell}(\mathbf{A})$.

\subsection{ViT as a Hybrid Model}

JEM reinterprets the standard softmax classifier as an EBM and trains a single network for hybrid discriminative-generative modeling. Specifically, JEM maximizes the logarithm of joint density function $p_{\bs{\theta}}(\bs{x},y)$:
\begin{equation}\label{eq:jem_loss}
  \log p_{\bs{\theta}}(\bs{x}, y) = \log p_{\bs{\theta}}(y|\bs{x}) + \log p_{\bs{\theta}}(\bs{x}),
\end{equation}
where the first term is the cross-entropy classification objective, and the second term can be optimized by the maximum likelihood learning of EBM using contrastive divergence and MCMC sampling. However, MCMC-based EBM is notorious due to the expensive $K$-step MCMC sampling that requires $K$ full forward and backward propagations at every iteration. Hence, removing the MCMC sampling in training is a promising direction~\cite{nomcmc}. 

\begin{figure}[t]
    \centering
        \includegraphics[width=0.9\columnwidth]{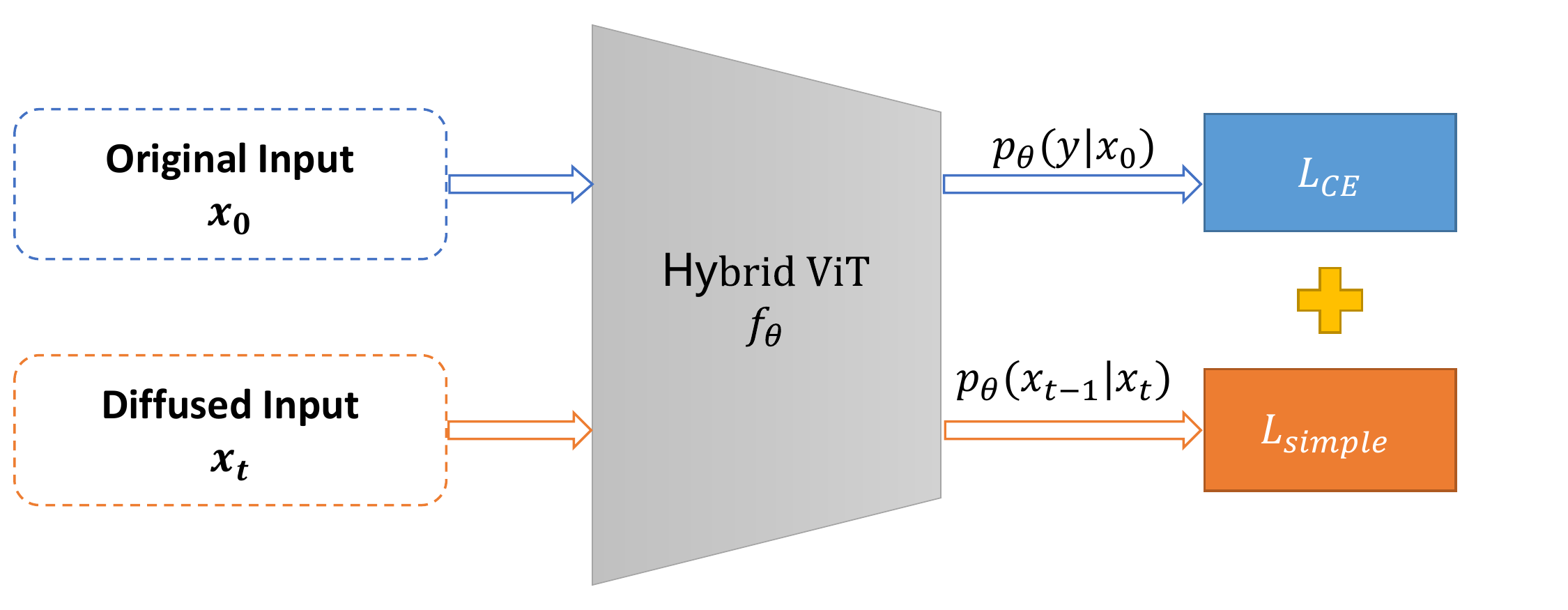}
    \caption{The pipeline of HybViT.}
    \label{fig:arch_hybvit}
\end{figure}

We propose Hybrid ViT (HybViT), a simple framework to extend GenViT for hybrid modeling. We substitute the optimization of $\log p_{\bs{\theta}}(\bs{x})$ in Eq.~\ref{eq:jem_loss} with the VLB of GenViT as Eq.~\ref{eq:elbo}.
Hence, we can train $p(y|x)$ using standard cross-entropy loss and optimize $p(x)$ using $L_{simple}$ loss in Eq~\ref{eq:loss_simple}. The final loss of our HybViT is 
\begin{align}
L = & L_{\text{CE}} + \alpha L_{\text{simple}}   \label{eq:total_loss}  \\
= & E_{\bs{x}_0, y} \left[ H(\bs{x}_0, y) \right] + \alpha E_{t,\vec{x}_0,\epsilon}\left[ || \epsilon - \epsilon_{\theta}(\vec{x}_t, t) ||^2 \right]   \label{eq:loss_eq}
\end{align}
We empirically find that a larger $\alpha=100$ improves the generation quality while retaining comparable classification accuracy. The training pipeline can be viewed in Fig~\ref{fig:arch_hybvit}.

\section{Experiments}

This section evaluates the discriminative and generative performance on multiple benchmark datasets, including CIFAR10, CIFAR100, STL10, CelebA-HQ-128, Tiny-ImageNet, and ImageNet 32x32. 

Our code is largely built on top of ViT~\cite{slvit2021}\footnote{\url{https://github.com/aanna0701/SPT_LSA_ViT}} and DDPM\footnote{\url{https://github.com/lucidrains/denoising-diffusion-pytorch}}. 
Note that we set the batch size as 128, and we update all ViT-based models with 1170 iterations in one epoch, while 390 iterations for CNN-based methods\footnote{ViT-based models use $3\times$ repeated augmentations~\cite{touvron2021training}}. Most experiments of ViTs run for 500 epochs, but 2500 epochs for STL10 and 100 epochs for ImageNet 32x32. Thanks to the memory efficiency of ViT, all our experiments can be performed with PyTorch on a single Nvidia GPU. For reproducibility, our source code is provided in the supplementary material.

\begin{figure}[ht]

    \centering
    \subfigure[CIFAR10]{
        \includegraphics[width=0.4\columnwidth]{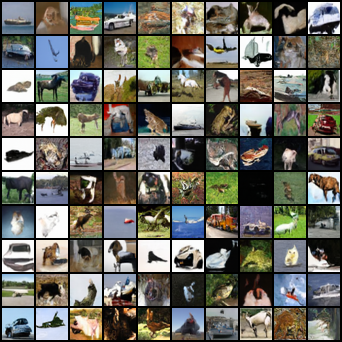}
        \label{figure:cifar10}
    }
    \subfigure[CelebA 128]{
        \includegraphics[width=0.4\columnwidth]{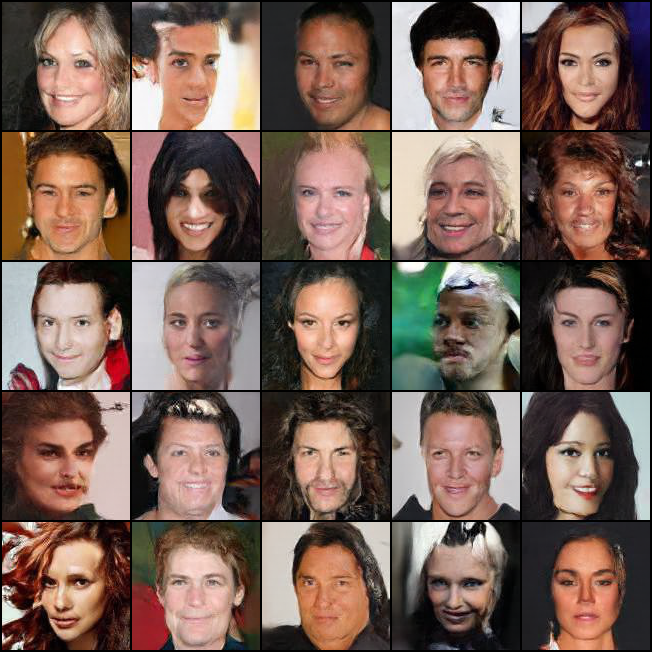}
        \label{figure:celeba}
    }
    \vspace{-10pt}
    \caption{GenViT Generated samples of CIFAR10 and CelebA 128.}
    \label{figure:genvit_cifar10_celeba}
    \vspace{-10pt}
\end{figure}

\subsection{Hybrid Modeling}

We first compare the performance with the state-of-the-art hybrid models, stand-alone discriminative and generative models on CIFAR10. We use accuracy, Inception Score (IS)~\cite{imprgan16} and Fr\'{e}chet Inception Distance (FID)~\cite{heusel2017gans} as evaluation metrics. IS and FID are employed to evaluate the quality of generated images. The results on CIFAR10 are shown in Tables~\ref{table:hybrid_cifar10}. HybViT outperforms other hybrid models including JEM ($K\!=\!20$) and JEM++ ($M\!=\!20$) on accuracy (95.9\%) and FID score (26.4), when the original ViT achieves comparable accuracy to WideResNet(WRN) 28-10. Moreover, GenViT and HybViT are superior in training stability. HybViT matches or outperforms the classification accuracy of JEM++ ($M\!=\!20$), and in the meantime, it exhibits high stability during training while JEM ($K\!=\!20$) and JEM++ ($M\!=\!5$) would easily diverge at early epochs. The comparison results on more benchmark datasets, including CIFAR100, STL10, CelebA-128, Tiny-ImageNet, ImageNet 32x32 are shown in Table~\ref{table:hybrid_results_other}. Example images generated by GenViT and HybViT are provided in Fig~\ref{figure:genvit_cifar10_celeba} and~\ref{figure:hybvit_cifar10_tinyimage}, respectively. More generated images can be found in the appendix.

\begin{table}[H]
\vspace{-10pt}
\caption{Results on CIFAR10.}
\vspace{-5pt}
\label{table:hybrid_cifar10}
\begin{center}
\begin{threeparttable}
\begin{tabular}{lccc}
\toprule
 Model                         & Acc \% $\uparrow$ & IS $\uparrow$ & FID $\downarrow$ \\
 \midrule
ViT                            & \bf{96.5} &  -   &  -   \\
GenViT                         &   -  & 8.17 & 20.2 \\
HybViT                         & 95.9 & 7.68 & 26.4 \\
\midrule
    \multicolumn{4}{c}{Single Hybrid Model} \\
IGEBM                          & 49.1 & 8.30 & 37.9 \\
JEM                            & 92.9 & 8.76 & 38.4 \\
JEM++ (M=20)                   & 94.1 & 8.11 & 38.0  \\
JEAT                           & 85.2 & 8.80 & 38.2  \\
\midrule 
\multicolumn{4}{c}{Generative Models} \\
SNGAN                          & - & 8.59  & 21.7  \\
StyleGAN2-ADA                  & - & \textbf{9.74}  & \textbf{2.92}  \\
DDPM                           & - & 9.46  & 3.17  \\ 
DiffuEBM                  & - &  8.31  & 9.58 \\
VAEBM                          & - &  8.43  & 12.2 \\
FlowEBM                        & - &   -    & 78.1 \\
\midrule 
\multicolumn{4}{c}{Other Models} \\
WRN-28-10                      & 96.2 & - & - \\
VERA(w/ generator)             & 93.2 & 8.11 & 30.5 \\
\bottomrule
\end{tabular}
\begin{tablenotes}
 \scriptsize\item  We compare with results reported by SNGAN~\cite{miyato2018spectral}, StyleGAN2-ADA~\cite{styleganADA}, DDPM~\cite{DDPM}, DiffuEBM~\cite{diffusionRecovery}, VAEBM~\cite{vaebm}, VERA~\cite{nomcmc}, FlowEBM~\cite{nijkamp2022mcmc}.
\end{tablenotes}
\end{threeparttable}
\vspace{-5pt}
\end{center}
\end{table}

\begin{figure}[ht]
 \vspace{-10pt}
    \centering
    \subfigure[CIFAR10]{
        \includegraphics[width=0.4\columnwidth]{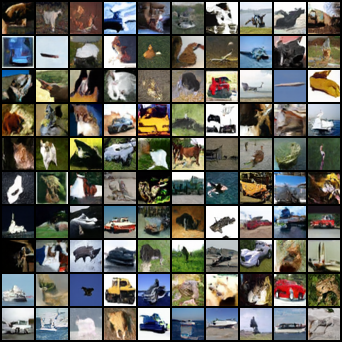}
        \label{figure:cifar10_hybrid}
    }
    \subfigure[STL10]{
        \includegraphics[width=0.4\columnwidth]{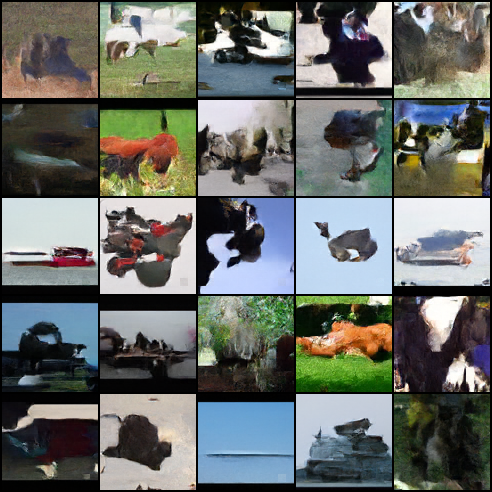}
        \label{figure:stl10}
    }
    \caption{HybViT Generated samples of CIFAR10 and STL10.}
    \vspace{-8pt}
    \label{figure:hybvit_cifar10_tinyimage}
\end{figure}

It's worth mentioning that the overall quality of synthesis is worse than UNet-based DDPM. In particular, our methods don't generate realistic images for complex and high-resolution data. ViT is known to model global relations between patches and lack of local inductive bias. We hope advances in ViT architectures and DDPM may address these issues in future work, such as Performer~\cite{performer_choromanski2021rethinking}, Swin Transformer~\cite{swintransformer}, CvT~\cite{wu2021cvt} and Analytic-DPM~\cite{bao2022analyticdpm}.

\begin{table}[H]
\vspace{-10pt}
\caption{Results on STL10, CelebA 128, Tiny-ImageNet, and ImageNet 32x32. Baseline models are selected based on availability in the literature.}
\vspace{-10pt}
\label{table:hybrid_results_other}
\begin{center}
\begin{threeparttable}
\begin{tabular}{lccc}
\toprule
 Model                         & Acc \% $\uparrow$ & IS $\uparrow$ & FID $\downarrow$ \\
 \midrule
\multicolumn{4}{c}{CIFAR100} \\
ViT                            & 77.8  &  -    &  -    \\
GenViT                         & -     & 8.19  & 26.0  \\
HybViT                         & 77.4  & 7.45  & 33.6  \\
WRN-28-10                      & \textbf{79.9}  &  -    &  -    \\
SNGAN                          & -     & 9.30  & 15.6  \\
BigGAN                         & -     & \textbf{11.0}  & \textbf{11.7} \\
 \midrule
\multicolumn{4}{c}{Tiny-ImageNet} \\
ViT                            & \textbf{57.6}  & - & - \\
GenViT                         &  -    & \textbf{7.81} & 66.7  \\
HybViT                         & 56.7  & 6.79 & 74.8  \\
PreactResNet18                 & 55.5  &  -   &  -    \\
ADC-GAN                        &  -    &  -   & \textbf{19.2}  \\
 \midrule
\multicolumn{4}{c}{STL10} \\
ViT                            & \textbf{84.2}  &  -   &  -    \\
GenViT                         & -     & 7.92 & 110   \\
HybViT                         & 80.8  & 7.87 & 109   \\
WRN-16-8                       & 76.6  &  -   & -     \\
SNGAN                          &   -   & \textbf{9.10} & \textbf{40.1}  \\
 \midrule
\multicolumn{4}{c}{ImageNet 32x32} \\
ViT                            & 57.5  &  -   &  -    \\
GenViT                         &  -    & 7.37 & 41.3  \\
HybViT                         & 53.5  & 6.66 & 46.4  \\
WRN-28-10                      & \textbf{59.1}  &  -   &  -    \\
IGEBM                          &   -   & 5.85 & 62.2  \\
KL-EBM                         &   -   & \textbf{8.73} & \textbf{32.4}  \\
 \midrule
\multicolumn{4}{c}{CelebA 128} \\
GenViT                         &  -    & -    &  22.07    \\
KL-EBM                         &  -    & -    &  28.78    \\  
SNGAN                          &  -    & -    &  24.36    \\
UNet GAN                      &  -    & -    &  \bf{2.95}     \\
\bottomrule
\end{tabular}
\begin{tablenotes}
 \scriptsize\item IGEBM~\cite{du2019implicit}, KL-EBM~\cite{improvedCD}, SNGAN~\cite{miyato2018spectral}, BigGAN~\cite{biggan}, ADC-GAN~\cite{adcgan}, UNet GAN~\cite{unetgan_schonfeld2020u}, .
\end{tablenotes}
\end{threeparttable}
\vspace{-10pt}
\end{center}
\end{table}

\subsection{Model Evaluation}

In this section, we conduct a thorough evaluation of proposed methods beyond the accuracy and generation quality. Note that it is not our intention to propose approaches to match or outperform the best models in all metrics.

\subsubsection{Calibration}

\begin{figure}[ht!]
\vspace{-5pt}
    \centering
    \subfigure[WRN]{
        \includegraphics[width=0.3\columnwidth]{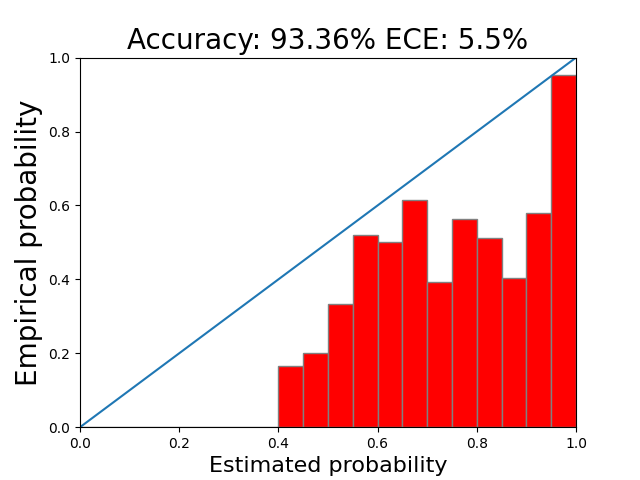}
        \label{figure:wrn_cali}
    }\vspace{-3pt}
    \subfigure[ViT]{
        \includegraphics[width=0.3\columnwidth]{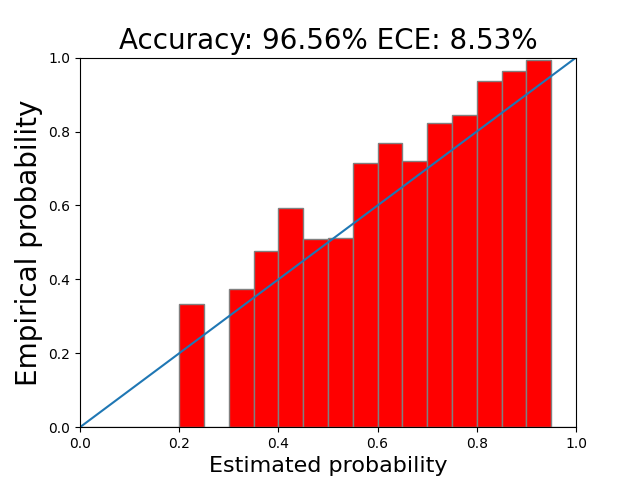}
        \label{figure:vit_cali}
    }\vspace{-3pt}
    \subfigure[HybViT]{
        \includegraphics[width=0.3\columnwidth]{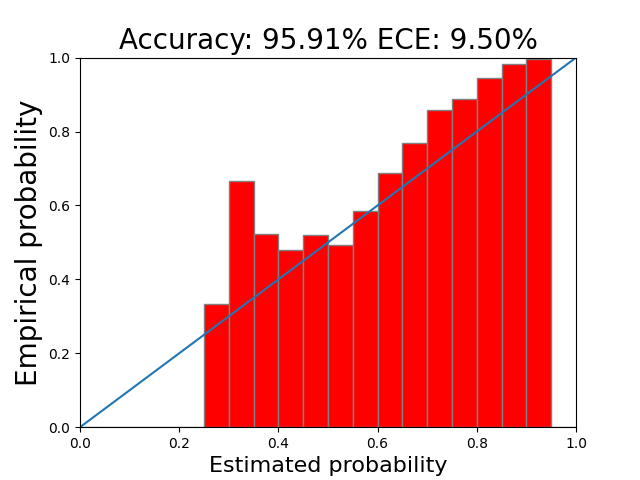}
        \label{figure:hybvit_cali}
    }\vspace{-3pt}
    
    \caption{Calibration results on CIFAR10. The smaller ECE is, the better. However, ViTs are better calibrated.}
    \label{figure:CIFAR10_cali}
    \vspace{-10pt}
\end{figure}

Recent works show that the predictions of modern convolutional neural networks could be over-confident due to increased model capacity~\cite{guo2017calibration}. Incorrect but confident predictions can be catastrophic for safety-critical applications. Hence, we investigate ViT and HybViT in terms of calibration using the metric Expected Calibration Error (ECE). Interestingly, Fig~\ref{figure:CIFAR10_cali} shows that predictions of both HybViT and ViT look like well-calibrated when trained with strong augmentations, however they are less confident and have worse ECE compared to WRN. More comparison results can be found in the appendix.

\subsubsection{Out-of-Distribution Detection}

Determining whether inputs are out-of-distribution (OOD) is an essential building block for safely deploying machine learning models in the open world. The model should be able to assign lower scores to OOD examples than to in-distribution examples such that it can be used to distinguish OOD examples from in-distribution ones. For evaluating the performance of OOD detection, we use a threshold-free metric, called Area Under the Receiver-Operating Curve (AUROC)~\cite{HenGim16}. Using the input density $p_{\bs{\theta}}(\bs{x})$~\cite{nalisnick2018deep} as the score, ViTs performs better in distinguishing the in-distribution samples from out-of-distribution samples as shown in Table~\ref{table:CIFAR10_ood},. 

\begin{table}[ht!]
\vspace{-5pt}
\caption{OOD detection results. Models are trained on CIFAR10 and values are AUROC.}
\label{table:CIFAR10_ood}
\begin{center}
\vspace{-8pt}
\begin{threeparttable}
\begin{tabular}{c|c|cccc}
\toprule
$s_{\bs{\theta}}(\bs{x})$  & Model   & SVHN & Interp & C100 & CelebA \\
\midrule
\multirow{8}{*}{$\log p_{\bs{\theta}}(\bs{x})$} & WRN* & .91 & - & \bf{.87} & .78 \\
            & IGEBM          & .63 & .70 & .50 & .70 \\
            & JEM            & .67 & .65 & .67 & .75 \\
            & JEM++          & .85 & .57 & .68 & .80 \\
            & VERA           & .83 & .86 & .73 & .33 \\
            & KL-EBM         & .91 & .65 & .83 & - \\
            & ViT            & \bf{.93} & \bf{.93} & .82 & \bf{.81} \\
            & HybViT         & \bf{.93} & .92 & .84 & .76\\
\bottomrule
\end{tabular}
\begin{tablenotes}
  \scriptsize\item * The result is from \cite{liu2020energy_ood}.
\end{tablenotes}
\end{threeparttable}
\vspace{-5pt}
\end{center}
\vspace{-5pt}
\end{table}

\subsubsection{Robustness}

\begin{figure}[H]
\vspace{-5pt}
    \centering
    \caption{Adversarial robustness under the PGD attacks.}
    \subfigure[$L_\infty$ Robustness]{
        \includegraphics[width=0.3\columnwidth]{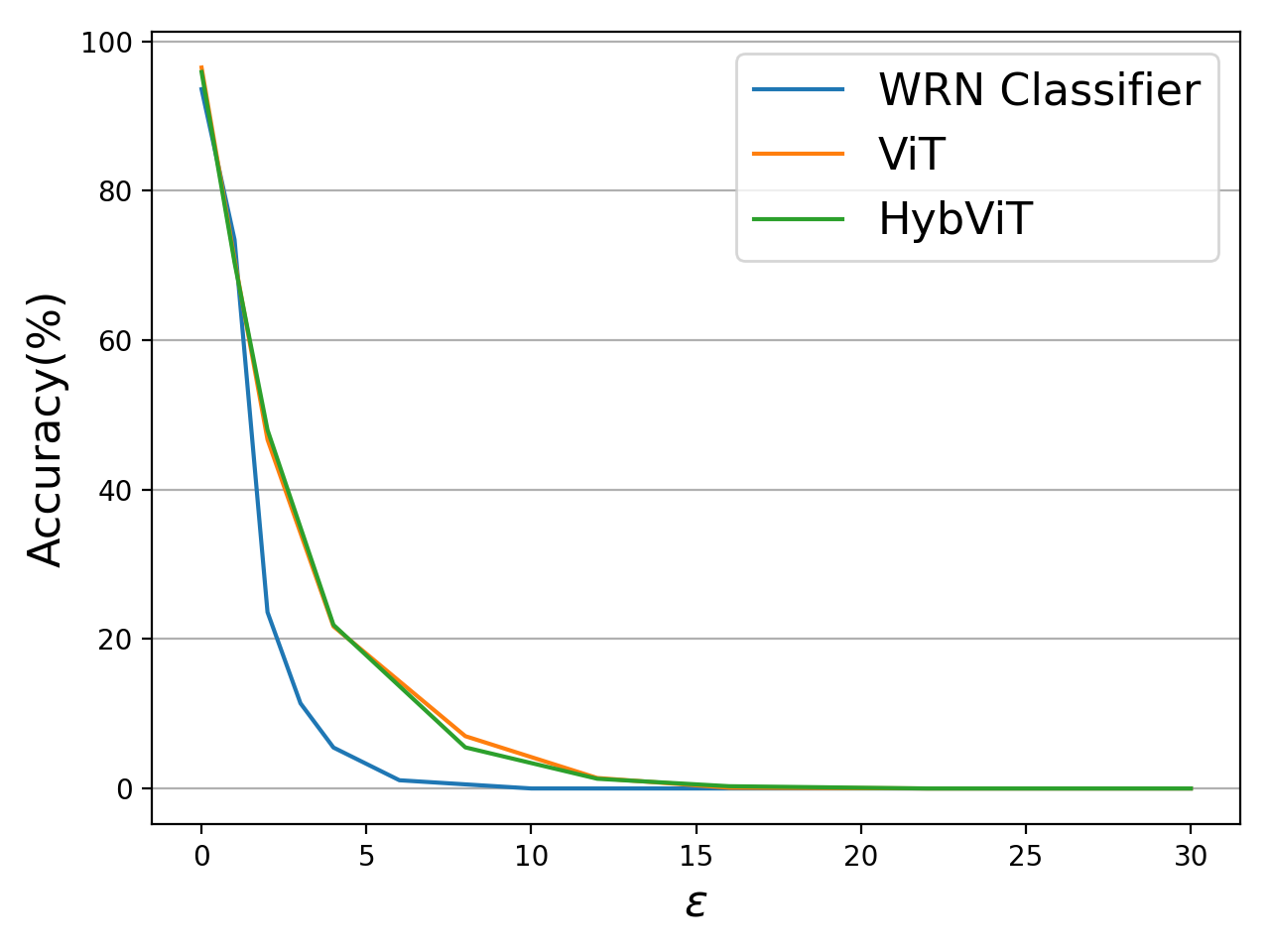} \label{figure:l_inf_robust}
    }\vspace{-5pt}
    \subfigure[$L_2$ Robustness]{
        \includegraphics[width=0.3\columnwidth]{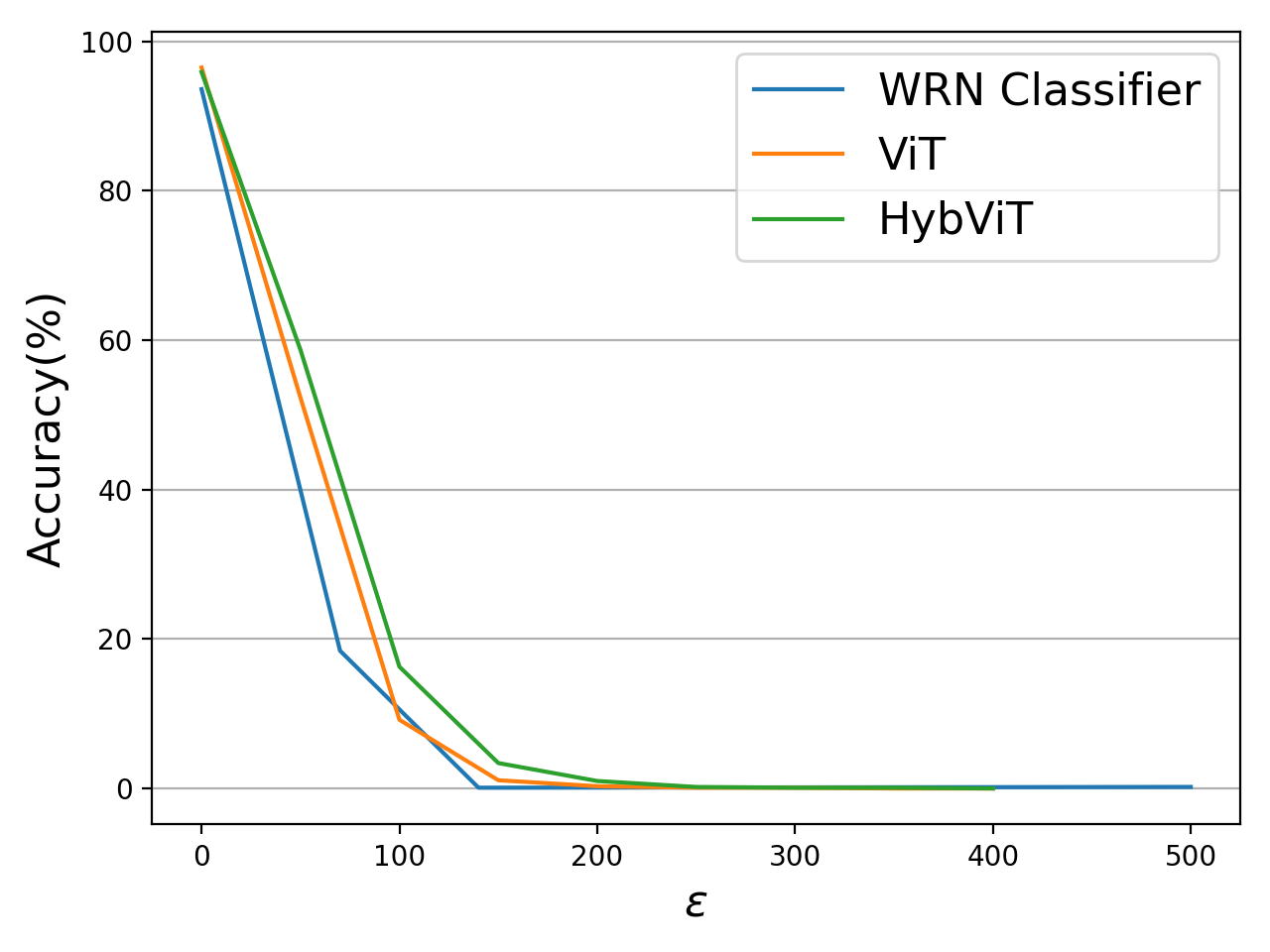} \label{figure:l2_robust}
    }\vspace{-5pt}
    \label{figure:robust_curve}

\end{figure}

Adversarial examples~\cite{propertiesNN14, goodfellow2014explaining} tricks the neural networks into giving incorrect predictions by applying minimal perturbations to the inputs, and hence, adversarial robustness is a critical characteristics of the model, which has received an influx of research interest. In this paper, we investigate the robustness of models trained on CIFAR10 using the white-box PGD attack~\cite{madry2018towards} under an $L_\infty$ or $L_2$ constraint. Fig~\ref{figure:robust_curve} compares ViT and HybViT with the baseline WRN-based classifier. We can see that ViT and HybViT have similar performance and both outperform WRN-based classifiers.  

\subsubsection{Likelihood}

An advantage of DDPM is that it can use the VLB as the approximated likelihood while most EBMs can't compute the intractable partition function w.r.t $\bs{x}$. Table~\ref{table:bpd} reports the test negative log-likelihood(NLL) in bits per dimension on CIFAR10. As we can observe, HybViT achieves comparable result to GenViT, and both are worse than other methods. 

\begin{table}[H]
\vspace{-5px}
\begin{center}

	\caption{NLL measured in bits/dim on CIFAR-10.} 
	\vspace{-5px}
 \begin{tabular}{lc} 
    \toprule
    Model                                   & BPD$\downarrow$ \\
    \midrule
    GenViT                                  & 3.78 \\
    HybViT                                  & 3.84 \\
    DDPM~\citep{DDPM}                       & 3.70 \\
    iDDPM~\citep{nichol2021iddpm}           & 2.94 \\
    DiffuEBM\citep{diffusionRecovery}  & 3.18 \\
    DistAug~\citep{da_gen2020jun}           & {\bf 2.56} \\
    \bottomrule 
    \end{tabular}
    \vspace{-5px}
    \label{table:bpd}
\end{center}
\vspace{-5px}
\end{table}

\subsection{Ablation Study}

In this section, we study the effect of different training configurations on the performance of image classification and generation by conducting an exhaustive ablation study on CIFAR10. We investigate the impact of 1) training epochs, 2) the coefficient $\alpha$, and 3) configurations of ViT/HybViT architecture in the main content. Due to page limitations, more results can be found in the appendix.

\begin{table}[H]
\vspace{-5pt}
\caption{Ablation study of epochs.}
\vspace{-5pt}
\label{table:epoch_factor}
\begin{center}
\vspace{-5pt}
\begin{threeparttable}
\begin{tabular}{lccc}
\toprule
 Model                  & Acc \% $\uparrow$ & IS $\uparrow$ & FID $\downarrow$ \\
 \midrule
ViT   (epoch=100)        & 94.2 &  -   &  -   \\
ViT   (epoch=300)        & 96.2 &  -   &  -   \\
ViT   (epoch=500)        & \bf{96.5} &  -   &  -   \\
GenViT(epoch=100)        &   -  & 7.25 & 33.3 \\
GenViT(epoch=300)        &   -  & 7.67 & 26.2 \\
GenViT(epoch=500)        &   -  & \bf{8.17} & \bf{20.2} \\
HybViT(epoch=100)        & 93.1 & 7.15 & 35.0 \\
HybViT(epoch=300)        & 95.9 & 7.59 & 29.5 \\
HybViT(epoch=500)        & 95.9 & 7.68 & 26.4 \\
\bottomrule
\end{tabular}
\end{threeparttable}
\vspace{-5pt}
\end{center}
\vspace{-5pt}
\end{table}

The results are reported in Table~\ref{table:epoch_factor} and  \ref{table:ablation_study}. First, Table~\ref{table:epoch_factor} shows a trade-off between the overall performance and computation time. The gain of classification and generation is relatively large when we prolong the training from 100 epochs to 300. With more training epochs, the accuracy gap between ViT and HybViT decreases. Furthermore, The generation quality can slightly improve after 300 epochs. 
Then we thoroughly explore the settings of the backbone ViT for GenViT and HybViT in Table~\ref{table:ablation_study}. It can be observed that larger $\alpha$ is preferred with high-quality generation and only small drop in accuracy. The number of heads also has a small effect on the trade-off between classification accuracy and generation quality. Enlarging the model capacity, depth, or hidden dimensions can improve the accuracy and generation quality.

\begin{table}[ht]
\vspace{-3pt}
\caption{Ablation Study on CIFAR10. The configurations of baseline of HybViT is $\alpha=100$,head=12,depth=9,dim=384.}
\vspace{-10pt}
\label{table:ablation_study}
\begin{center}
\begin{threeparttable}
\begin{tabular}{lccc}
\toprule
 Model                   & Acc \% $\uparrow$ & IS $\uparrow$ & FID $\downarrow$ \\
\midrule
HybViT                   & 95.9 & 7.68  & 26.4 \\
 \midrule
HybViT($\alpha$=1)       & 96.6 & 4.74  & 68.9 \\
HybViT($\alpha$=10)      & \bf{97.0} & 6.40  & 38.2 \\
\midrule 
HybViT(head=6)           & 96.0 & 7.51  & 30.0 \\
HybViT(head=8)           & 95.9 & 7.74  & 28.0 \\
HybViT(head=16)          & 95.4 & 7.79  & 27.1 \\
\midrule 
HybViT(depth=6)          & 94.7 & 7.39  & 30.6 \\
HybViT(depth=12)         & 96.6 & 7.78  & 24.3 \\
\midrule 
HybViT(dim=192)          & 94.1 & 7.06  & 35.0 \\
HybViT(dim=768)          & 96.4 & 8.04  & 19.9 \\
\midrule 
GenViT(dim=192)          &   -  & 7.26  & 32.5 \\
GenViT(dim=384)          &   -  & 8.17  & 20.2 \\
GenViT(dim=768)          &   -  & \bf{8.32}  & \bf{18.7} \\
\bottomrule
\end{tabular}
\end{threeparttable}
\end{center}
\end{table}

While it is challenging for our methods to generate realistic images for complex and high-resolution data, it is beyond the scope of this work to further improve the generation quality for high-resolution data. Thus, it warrants an exciting direction of future work. We suppose the large patch size of the ViT's architecture is the critical causing factor. Hence, we investigate the impact of different patch sizes on STL10 in Table~\ref{table:ablation_study_stl10}. However, even though a smaller patch size can improve the accuracy by a notably margin at the cost of increasing model sizes, but the generation quality for high-resolution images plateaued around $p=6$.
These results indicate that the bottleneck of image generation comes from other components, such as the linear projections and reconstruction projections, other than the multi-head self-attention. Note that a larger patch size (ps=12) do further deteriorate the generation quality and would lead to critical issues for high-resolution data like ImageNet, since the corresponding patch size is typically set to 14 or larger.

\begin{table}[ht]
\caption{Ablation Study on STL10. All models are trained for 500 epochs. NoP means Number of Parameters. ps means Patch Size.}
\vspace{-8pt}
\label{table:ablation_study_stl10}
\begin{center}
\begin{threeparttable}
\begin{tabular}{lcccc}
\toprule
 Model            & NoP    & Acc \% $\uparrow$ & IS $\uparrow$ & FID $\downarrow$ \\
 \midrule
ViT(ps=8)         & 12.9M  & 78.7    &   -   & - \\
HybViT(ps=4)      & 41.1M  & \bf{87.1}    & 6.90  & 125.5 \\
HybViT(ps=6)      & 17.0M  & 81.7    & \bf{7.30}  & \bf{123.6} \\
HybViT(ps=8)      & 12.9M  & 77.5    & 6.95  & 125.2 \\
HybViT(ps=12)     & 11.4M  & 69.1    & 2.55  & 240.2 \\
\midrule
GenViT(dim=384)   & 12.9M  &  -      & 6.95  & 125.2 \\
GenViT(dim=576)   & 26.4M  &  -      & 7.02  & 124.1 \\
GenViT(dim=768)   & 45.2M  &  -      & 7.01  & 126.6 \\
\bottomrule
\end{tabular}
\vspace{-5pt}
\end{threeparttable}
\vspace{-3pt}
\end{center}
\vspace{-5pt}
\end{table}

\subsubsection{Training Speed}

We report the empirical training speeds of our models and baseline methods on a single GPU for CIFAR10 in Table~\ref{table:speed} and those for ImageNet 32x32 is in the appendix. As discussed previously, two mini-batches are utilized in HybViT: one for training of $L_{simple}$ and the other for training of the cross entropy loss. 
Hence, HybViT requires about $2\times$ training time compared to GenViT. 
One of the advantages of GenViT and HybViT is that even with much more ($7.5\times$) iterations, they still reduce training time significantly compared to EBMs. 
The results demonstrate that our new methods are much faster and affordable for academia research settings.

\begin{table}[ht]
\vspace{-5pt}
\caption{Run-time comparison on CIFAR10. We set 1170 iterations as one epoch for all ViTs, and 390 for WRN-based models. All ViTs are trained for 500 epochs and WRN-based models are trained for 200 epochs.}
\vspace{-10pt}

\label{table:speed}
\begin{center}

\begin{threeparttable}
\begin{tabular}{lccc}
\toprule
Model              &  NoP(M) & Min/Epoch  & Runtime(Hours)  \\
\midrule
\multicolumn{4}{c}{ViT-based Models}   \\
ViT(d=384)         &  11.2 & 1.72  & 14.4  \\
GenViT(d=384)      &  11.2 & 2.11  & 17.6  \\
HybViT(d=192)      &  3.2  & 2.14  & 17.9  \\
HybViT(d=384)      &  11.2 & 3.71  & 31.2  \\
HybViT(d=768)      &  43.2 & 9.34  & 77.8  \\
\midrule
\multicolumn{4}{c}{WRN-based Models}   \\
WRN 28-10          &  36.5 & 1.53  & 5.2   \\
JEM(K=20)          &  36.5 & 30.2  & 101.3 \\
JEM++(K=10)        &  36.5 & 20.4  & 67.4  \\
VERA               &  40 &  19.3   &  64.3  \\
\midrule
IGEBM              & -     &  \multicolumn{2}{c}{1 GPU for 2 days}              \\
KL-EBM           &  6.64 &  \multicolumn{2}{c}{1 GPU for 1 day}        \\
VAEBM*             & 135 &  \multicolumn{2}{c}{400 epochs, 8 GPUs, 55 hours}  \\
DDPM               &  35.7 &  \multicolumn{2}{c}{800k iter, 8 TPUs, 10.6 hours} \\
DiffuEBM         & 34.8  &  \multicolumn{2}{c}{240k iter, 8 TPUs, 40+ hours}  \\
\bottomrule
\end{tabular}
\begin{tablenotes}
  \scriptsize\item * The runtime is for pretraining NVAE only. It further needs 25,000 iterations (or 16 epochs) on CIFAR-10 using one GPU for VAEBM.
\end{tablenotes}
\end{threeparttable}
\end{center}
\end{table}

\section{Limitations}

As shown in previous sections, our models GenViT and HybViT exhibit promising results. However, compared to CNN-based methods, the main limitations are: 1) The generation quality is relatively low compared with pure generation (non-hybrid) SOTA models. 2) They require more training iterations to achieve high classification performance compared with pure classification models. 3) The sampling speed during inference is slow (typically $T \geq 1000$) while GAN only needs one-time forward. 

We believe the results presented in this work are sufficient to motivate the community to solve these limitations and improve speed and generative quality.

\section{Conclusion}

In this work, we integrate a single ViT into DDPM to propose a new type of generative model, GenViT. Furthermore, we present HybViT, a simple approach for training hybrid discriminative-generative models. We conduct a series of thorough experiments to demonstrate the effectiveness of these models on multiple benchmark datasets with state-of-the-art results in most of the tasks of image classification, and image generation. We also investigate the intriguing properties, including likelihood, adversarial robustness, uncertainty calibration, and OOD detection. Most importantly, the proposed approach HybViT provides stable training, and outperforms the previous state-of-the-art hybrid models on both discriminative and generation tasks. While there are still challenges in training the models for high-resolution images, we hope the results presented here will encourage the community to improve upon current approaches.

\bibliographystyle{ieee_fullname}
\bibliography{main_bib}

\appendix

\section{Image Datasets}\label{app:datasets}
The image benchmark datasets used in our experiments are described below:
\begin{enumerate}
\item CIFAR10~\cite{Krizhevsky2012} contains 60,000 RGB images of size $32\times 32$ from 10 classes, in which 50,000 images are for training and 10,000 images are for test.
\item CIFAR100~\cite{Krizhevsky2012} also contains 60,000 RGB images of size $32\times 32$, except that it contains 100 classes with 500 training images and 100 test images per class.
\item STL10~\cite{stl10} 500 training images from 10 classes as CIFAR10, 800 test images per class.
\item Tiny-ImageNet contains 100000 images of 200 classes (500 for each class) downsized to 64×64 colored images. Each class has 500 training images, 50 validation images and 50 test images.
\item CelebA-HQ~\cite{karras2018progressive} is a human face image dataset. In our experiment, we use the downsampled version with size $128\times 128$.
\item Imagenet 32x32~\cite{imgnet32} is a downsampled variant of ImageNet with 1,000 classes. It contains the same number of images as vanilla ImageNet, but the image size is $32\times 32$.
\end{enumerate}

\section{Experimental Details}\label{app:exp}

As we discuss in the main content, all our experiments are based on vanilla ViT in ~\cite{slvit2021}\footnote{\url{https://github.com/aanna0701/SPT_LSA_ViT}} and DDPM\footnote{\url{https://github.com/lucidrains/denoising-diffusion-pytorch}} and follow their settings. We use SGD for all datasets with an initial learning rate of 0.1. We reduce the learning rate using the cosine scheduler. 
Table~\ref{table:hyperparameters_difvit_app} lists the hyper-parameters in our experiments. We also tried $T = 4000$ and $L_2$ loss to train our GenViT and HybViT, and their final results are comparable.

\begin{table}[ht!]
\caption{Hyper-parameters of ViT, GenViT and HybViT}
\label{table:hyperparameters_difvit_app}
\begin{center}
\begin{threeparttable}
\begin{tabular}{lc}
\toprule
Variable                         & Value      \\
\midrule
Learning rate                    & 0.1        \\
Batch Size                       & 128        \\
Warmup Epochs                    & 10         \\
Coefficient $\alpha$ in HybViT   & 1, 10, 100 \\
\midrule
\multicolumn{2}{c}{Configurations of ViT}     \\
Dimensions                       & 384        \\
Depth                            & 9          \\
Heads                            & 12         \\
Patch Size                       & 4, 8       \\
\midrule
\multicolumn{2}{c}{Configurations of DDPM}    \\
Number of Timesteps $T$          & 1000       \\
Loss Type                        & $L_1$      \\
Noise Schedule                   & cosine     \\
\bottomrule
\end{tabular}
\end{threeparttable}
\end{center}
\end{table}

\begin{figure}[H]

    \centering

    \subfigure[CIFAR10]{
        \includegraphics[width=0.45\columnwidth]{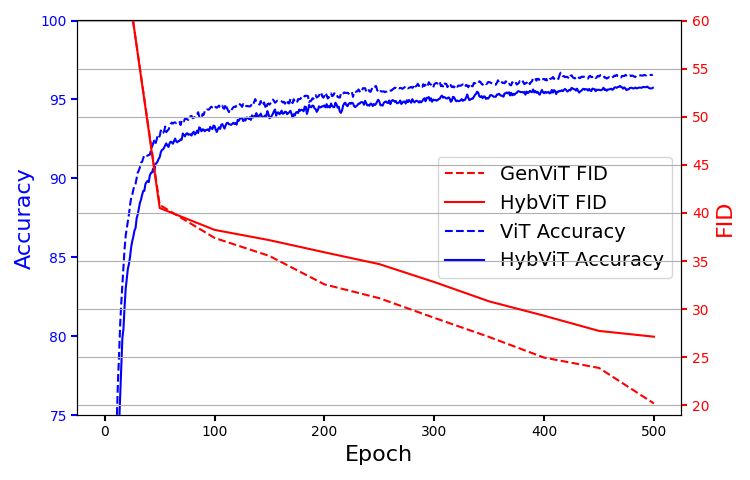}
        \label{figure:cifar10_metrics}
    }
    \subfigure[ImageNet 32x32]{
        \includegraphics[width=0.45\columnwidth]{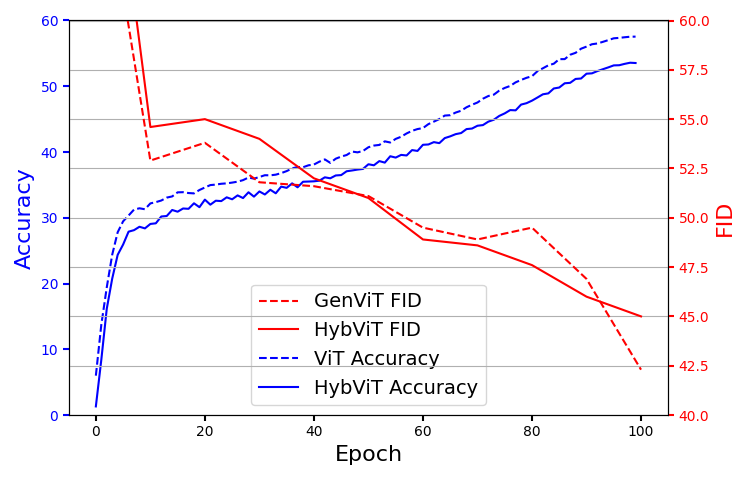}
        \label{figure:img32_metrics}
    }
    \caption{The evolution of HybViT's classification accuracy, FID as a function of training epochs on CIFAR10 and ImageNet 32x32.}
    \label{figure:evolution_acc_fid}

\end{figure}

\begin{figure}[ht]

    \centering

    \subfigure[Samples with FID=20]{
        \includegraphics[width=0.4\columnwidth]{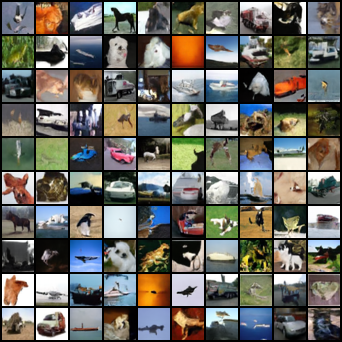}
        \label{figure:fid20_difvit_app}
    }
    \subfigure[Samples with FID=40]{
        \includegraphics[width=0.4\columnwidth]{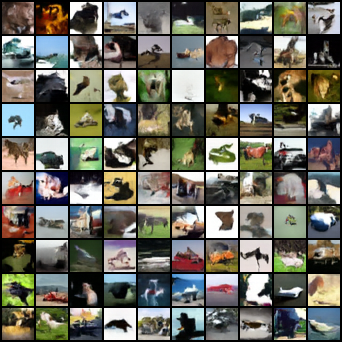}
        \label{figure:fid40_difvit_app}
    }
    \caption{The comparison between samples with FID=40 and FID=20. The difference is visually imperceptible for human.}
    \label{figure:cmp_fid_20_40}

\end{figure}

\section{Model Evaluation}

\subsection{Qualitative Analysis of Samples}

First, we investigate the gap between ViT, GenViT and HybViT in Fig~\ref{figure:evolution_acc_fid}. We select two benchmark datasets CIFAR10 and ImageNet 32x32. It can be observed that  the improvement of generation quality is relatively small after 10\% training epochs. The difference is almost visually imperceptible for human between samples with FID=40 and FID=20 as shown in Fig. Hence, we think accelerating the convergence rates of our models is an interesting direction in the future.

\begin{figure*}[ht]

    \centering

    \subfigure[Samples with highest $E(\bs{x})$]{
        \includegraphics[width=0.23\columnwidth]{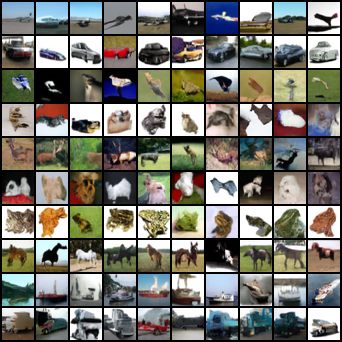} \label{figure:top_px}
    }
    \subfigure[Samples with lowest $E(\bs{x})$]{
        \includegraphics[width=0.23\columnwidth]{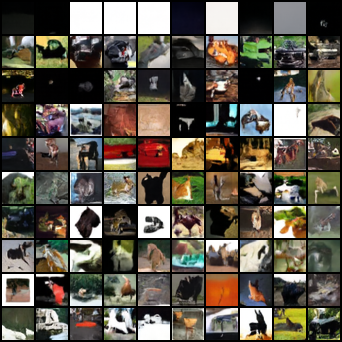} \label{figure:bottom_px}
    }
    \subfigure[Samples with highest $p(y|\bs{x})$]{
        \includegraphics[width=0.23\columnwidth]{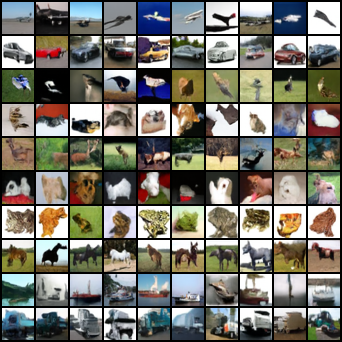} \label{figure:top_pyx}
    }
    \subfigure[Samples with lowest $p(y|\bs{x})$]{
        \includegraphics[width=0.23\columnwidth]{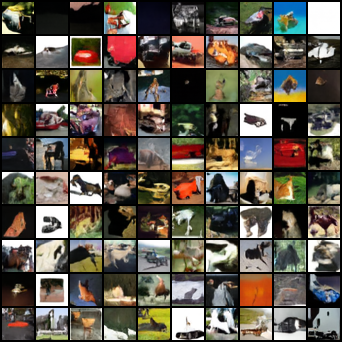} \label{figure:bottom_pyx}
    }
    \caption{HybViT generated class-conditional (best and worst) samples. Each row corresponds to 1 class.}
    \label{figure:categorial_topk_bottomk}

\end{figure*}

\begin{figure*}[ht]
    \centering
        \includegraphics[width=1\textwidth]{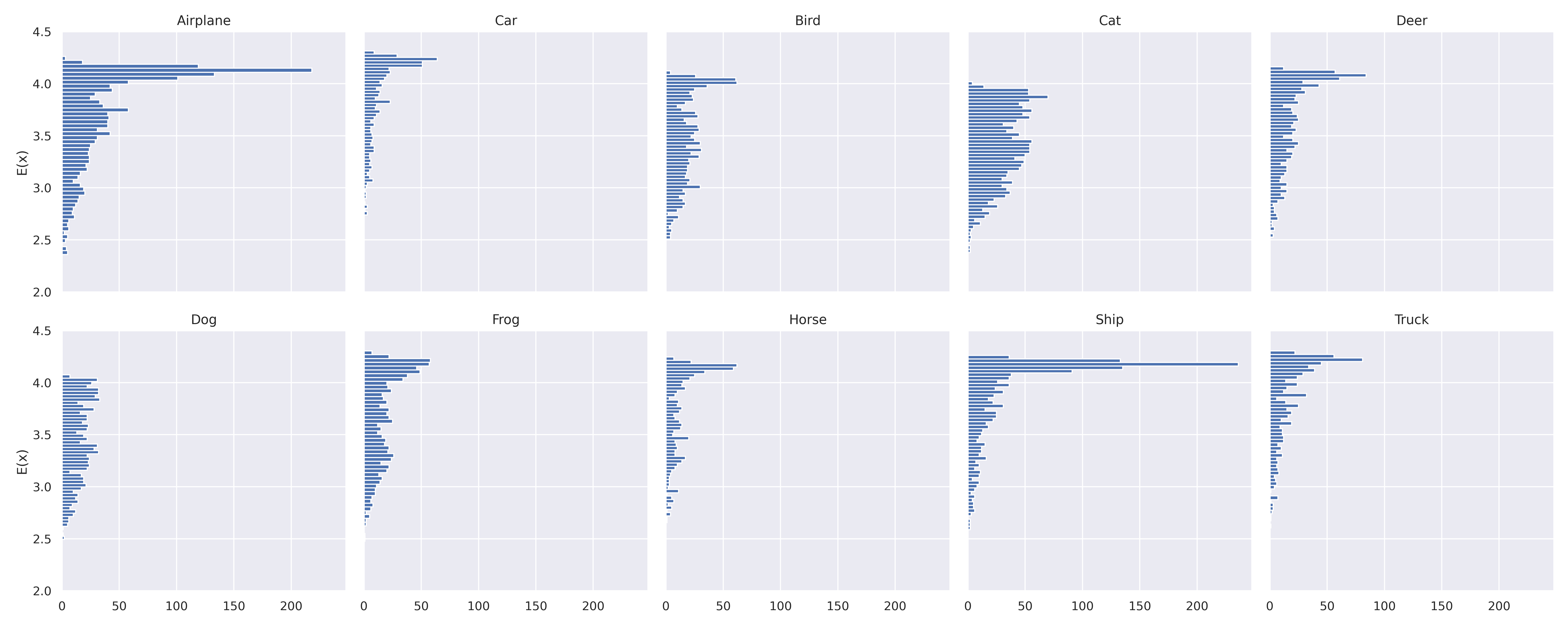}
    \caption{Histograms (oriented horizontally for easier visual alignment) of $\bs{x}$ arranged by class for CIFAR10.}
    \label{fig:px_category}
\end{figure*}

\begin{figure*}[ht!]
    \centering
    \subfigure[Samples with highest $p(\bs{x}$)]{
        \includegraphics[width=0.22\columnwidth]{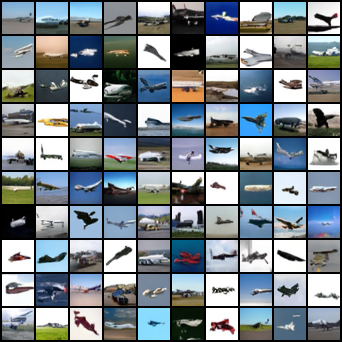}
    }
    \subfigure[Samples with lowest $p(\bs{x})$]{
        \includegraphics[width=0.22\columnwidth]{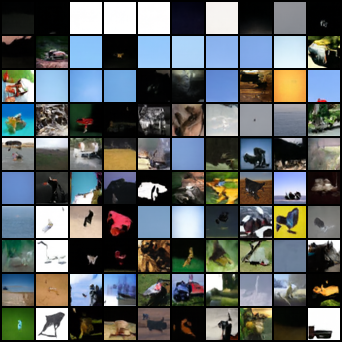}
    }
    \subfigure[Samples with highest $p(y|\bs{x}$)]{
        \includegraphics[width=0.22\columnwidth]{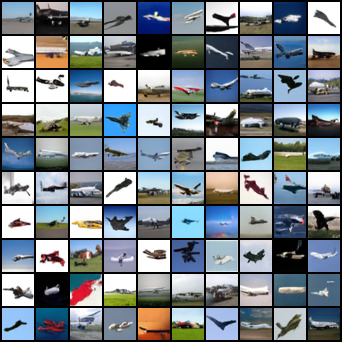}
    }
    \subfigure[Samples with lowest $p(y|\bs{x})$]{
        \includegraphics[width=0.22\columnwidth]{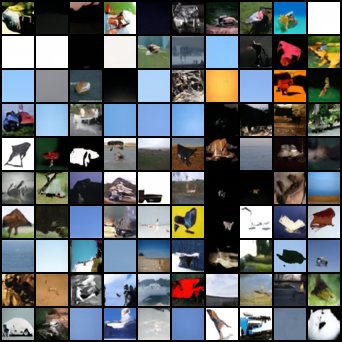}
    }
    \caption{HybViT generated class-conditional samples of \textbf{Plane}}
    \label{figure:hybvit_app_class_0}
\end{figure*}

\begin{figure*}[ht!]
    \centering
    \subfigure[Samples with highest $p(\bs{x}$)]{
        \includegraphics[width=0.22\columnwidth]{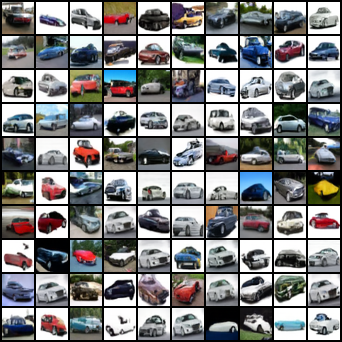}
    }
    \subfigure[Samples with lowest $p(\bs{x})$]{
        \includegraphics[width=0.22\columnwidth]{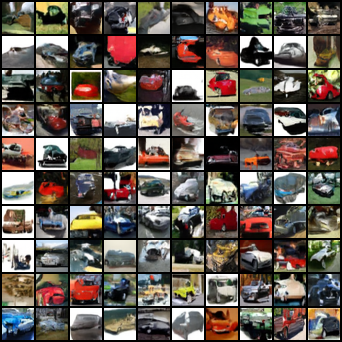}
    }
    \subfigure[Samples with highest $p(y|\bs{x}$)]{
        \includegraphics[width=0.22\columnwidth]{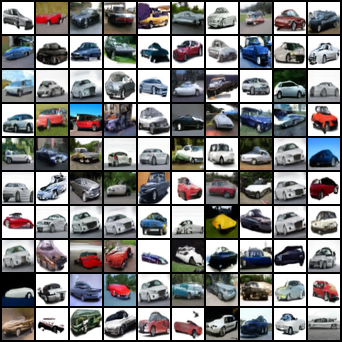}
    }
    \subfigure[Samples with lowest $p(y|\bs{x})$]{
        \includegraphics[width=0.22\columnwidth]{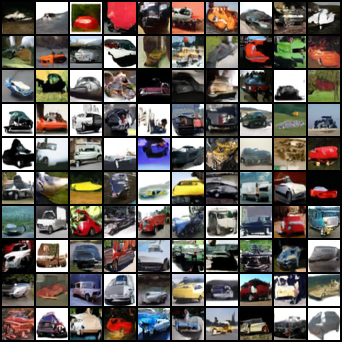}
    }
    \caption{HybViT generated class-conditional samples of \textbf{Car}}
    \label{figure:hybvit_app_class_1}
\end{figure*}

\begin{figure*}[ht!]
    \centering
    \subfigure[Samples with highest $p(\bs{x}$)]{
        \includegraphics[width=0.22\columnwidth]{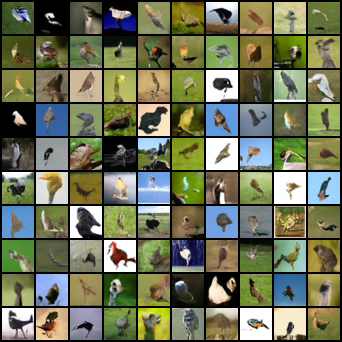}
    }
    \subfigure[Samples with lowest $p(\bs{x})$]{
        \includegraphics[width=0.22\columnwidth]{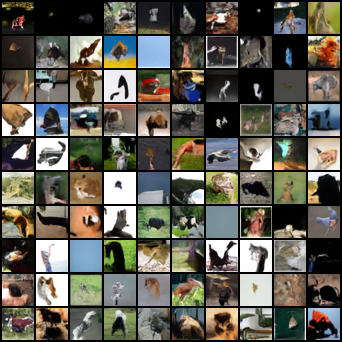}
    }
    \subfigure[Samples with highest $p(y|\bs{x}$)]{
        \includegraphics[width=0.22\columnwidth]{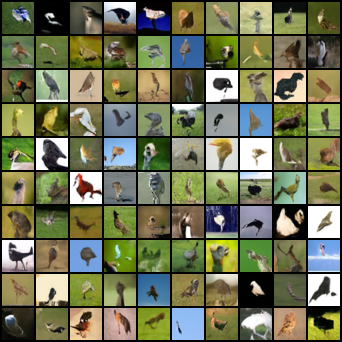}
    }
    \subfigure[Samples with lowest $p(y|\bs{x})$]{
        \includegraphics[width=0.22\columnwidth]{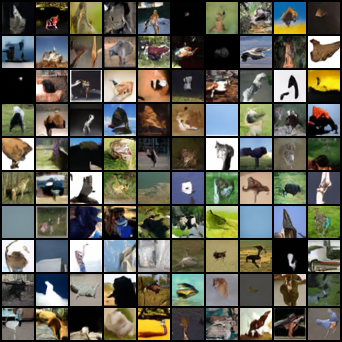}
    }
    \caption{HybViT generated class-conditional samples of \textbf{Bird}}
    \label{figure:hybvit_app_class_2}
\end{figure*}

\begin{figure*}[ht!]
    \centering
    \subfigure[Samples with highest $p(\bs{x}$)]{
        \includegraphics[width=0.22\columnwidth]{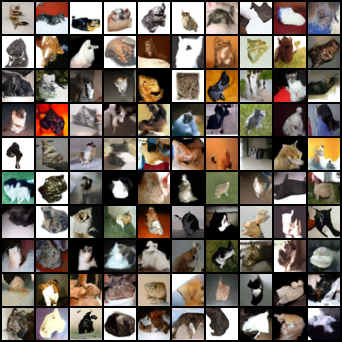}
    }
    \subfigure[Samples with lowest $p(\bs{x})$]{
        \includegraphics[width=0.22\columnwidth]{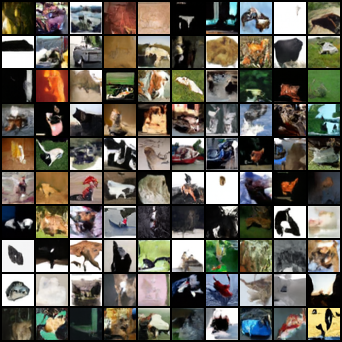}
    }
    \subfigure[Samples with highest $p(y|\bs{x}$)]{
        \includegraphics[width=0.22\columnwidth]{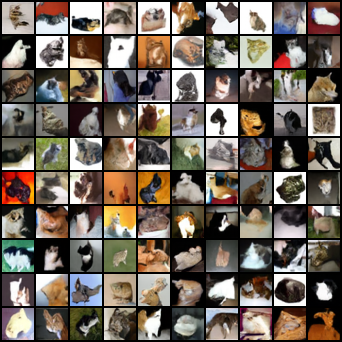}
    }
    \subfigure[Samples with lowest $p(y|\bs{x})$]{
        \includegraphics[width=0.22\columnwidth]{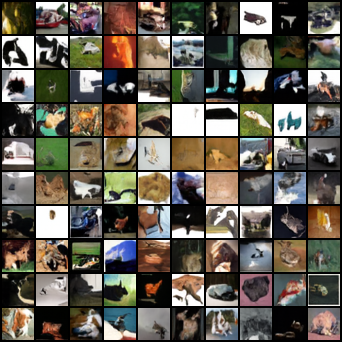}
    }
    \caption{HybViT generated class-conditional samples of \textbf{Cat}}
    \label{figure:hybvit_app_class_3}
\end{figure*}

\begin{figure*}[ht!]
    \centering
    \subfigure[Samples with highest $p(\bs{x}$)]{
        \includegraphics[width=0.22\columnwidth]{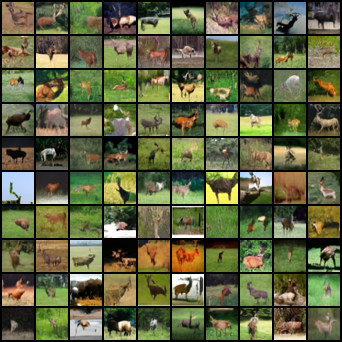}
    }
    \subfigure[Samples with lowest $p(\bs{x})$]{
        \includegraphics[width=0.22\columnwidth]{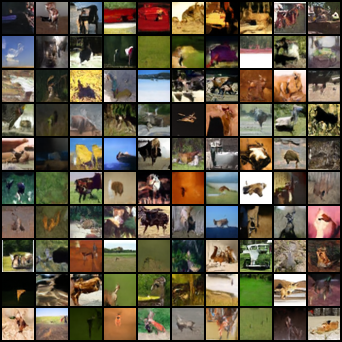}
    }
    \subfigure[Samples with highest $p(y|\bs{x}$)]{
        \includegraphics[width=0.22\columnwidth]{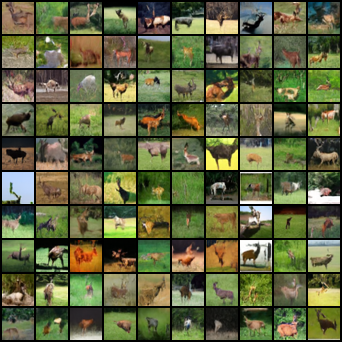}
    }
    \subfigure[Samples with lowest $p(y|\bs{x})$]{
        \includegraphics[width=0.22\columnwidth]{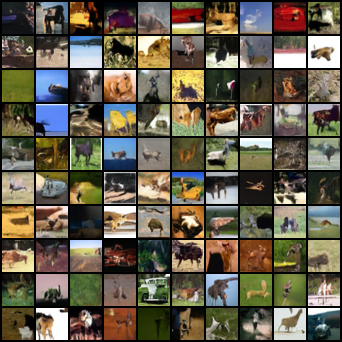}
    }
    \caption{HybViT generated class-conditional samples of \textbf{Deer}}
    \label{figure:hybvit_app_class_4}
\end{figure*}

\begin{figure*}[ht!]
    \centering
    \subfigure[Samples with highest $p(\bs{x}$)]{
        \includegraphics[width=0.22\columnwidth]{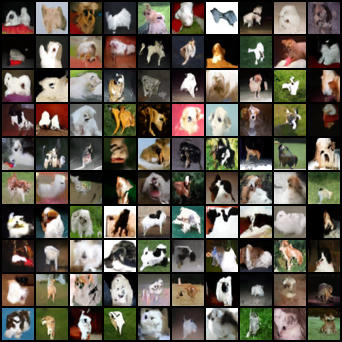}
    }
    \subfigure[Samples with lowest $p(\bs{x})$]{
        \includegraphics[width=0.22\columnwidth]{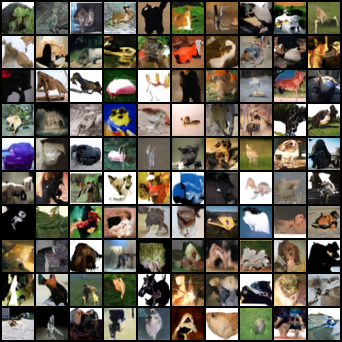}
    }
    \subfigure[Samples with highest $p(y|\bs{x}$)]{
        \includegraphics[width=0.22\columnwidth]{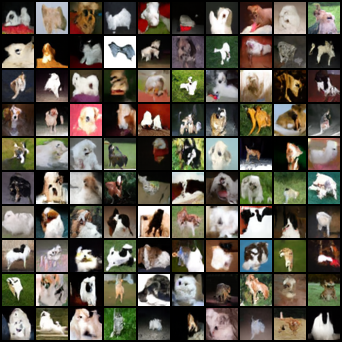}
    }
    \subfigure[Samples with lowest $p(y|\bs{x})$]{
        \includegraphics[width=0.22\columnwidth]{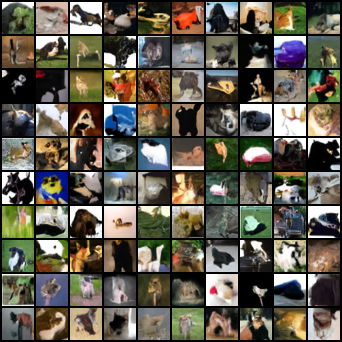}
    }
    \caption{HybViT generated class-conditional samples of \textbf{Dog}}
    \label{figure:hybvit_app_class_5}
\end{figure*}

\begin{figure*}[ht!]
    \centering
    \subfigure[Samples with highest $p(\bs{x}$)]{
        \includegraphics[width=0.22\columnwidth]{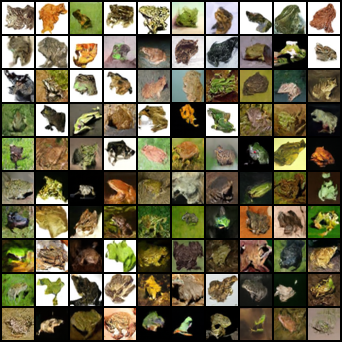}
    }
    \subfigure[Samples with lowest $p(\bs{x})$]{
        \includegraphics[width=0.22\columnwidth]{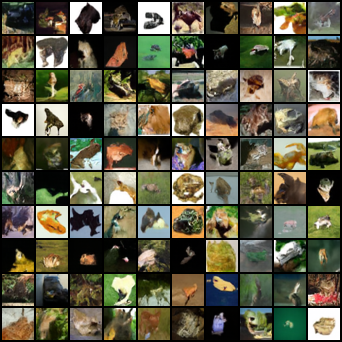}
    }
    \subfigure[Samples with highest $p(y|\bs{x}$)]{
        \includegraphics[width=0.22\columnwidth]{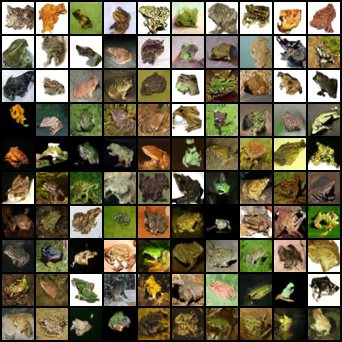}
    }
    \subfigure[Samples with lowest $p(y|\bs{x})$]{
        \includegraphics[width=0.22\columnwidth]{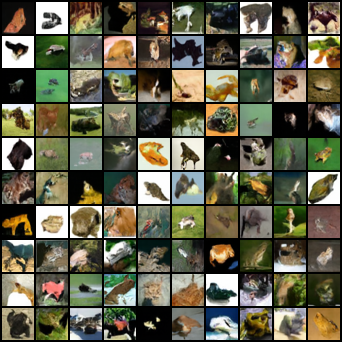}
    }
    \caption{HybViT generated class-conditional samples of \textbf{Frog}}
    \label{figure:hybvit_app_class_6}
\end{figure*}

\begin{figure*}[ht!]
    \centering
    \subfigure[Samples with highest $p(\bs{x}$)]{
        \includegraphics[width=0.22\columnwidth]{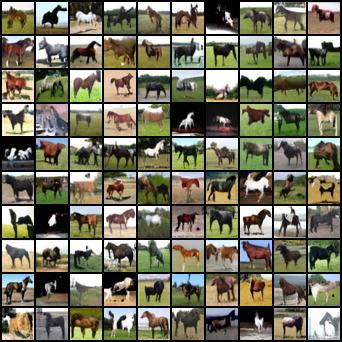}
    }
    \subfigure[Samples with lowest $p(\bs{x})$]{
        \includegraphics[width=0.22\columnwidth]{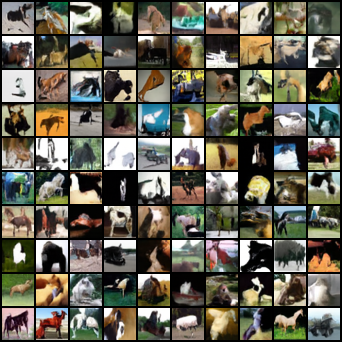}
    }
    \subfigure[Samples with highest $p(y|\bs{x}$)]{
        \includegraphics[width=0.22\columnwidth]{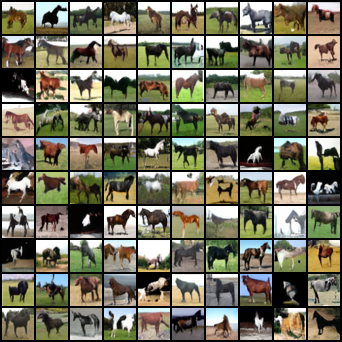}
    }
    \subfigure[Samples with lowest $p(y|\bs{x})$]{
        \includegraphics[width=0.22\columnwidth]{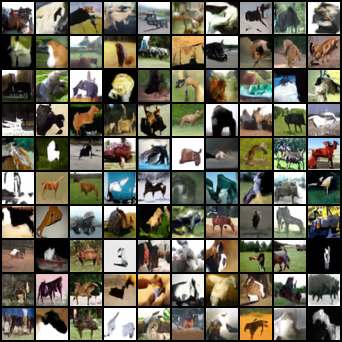}
    }
    \caption{HybViT generated class-conditional samples of \textbf{Horse}}
    \label{figure:hybvit_app_class_7}
\end{figure*}

\begin{figure*}[ht!]
    \centering
    \subfigure[Samples with highest $p(\bs{x}$)]{
        \includegraphics[width=0.22\columnwidth]{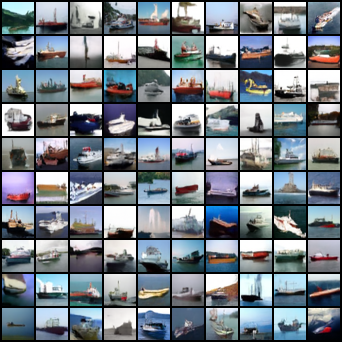}
    }
    \subfigure[Samples with lowest $p(\bs{x})$]{
        \includegraphics[width=0.22\columnwidth]{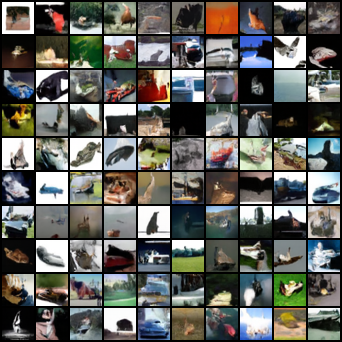}
    }
    \subfigure[Samples with highest $p(y|\bs{x}$)]{
        \includegraphics[width=0.22\columnwidth]{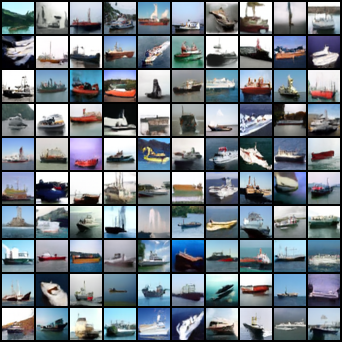}
    }
    \subfigure[Samples with lowest $p(y|\bs{x})$]{
        \includegraphics[width=0.22\columnwidth]{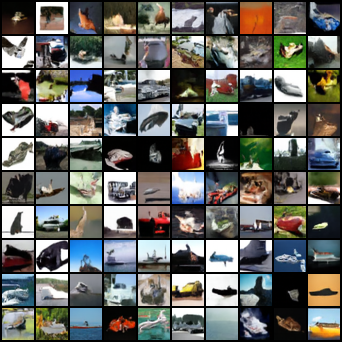}
    }
    \caption{HybViT generated class-conditional samples of \textbf{Ship}}
    \label{figure:hybvit_app_class_8}
\end{figure*}

\begin{figure*}[ht!]
    \centering
    \subfigure[Samples with highest $p(\bs{x}$)]{
        \includegraphics[width=0.22\columnwidth]{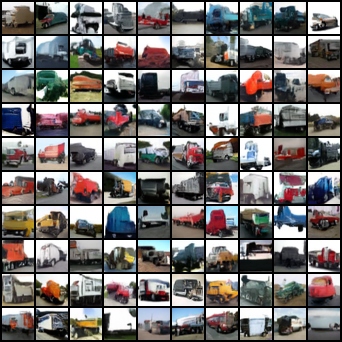}
    }
    \subfigure[Samples with lowest $p(\bs{x})$]{
        \includegraphics[width=0.22\columnwidth]{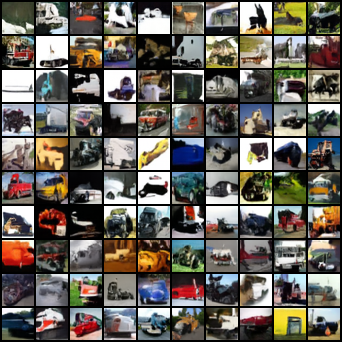}
    }
    \subfigure[Samples with highest $p(y|\bs{x}$)]{
        \includegraphics[width=0.22\columnwidth]{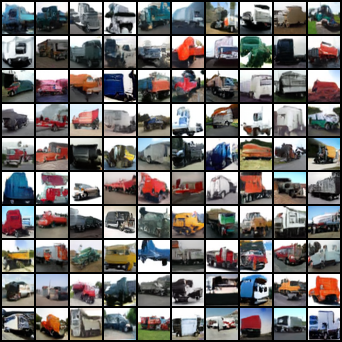}
    }
    \subfigure[Samples with lowest $p(y|\bs{x})$]{
        \includegraphics[width=0.22\columnwidth]{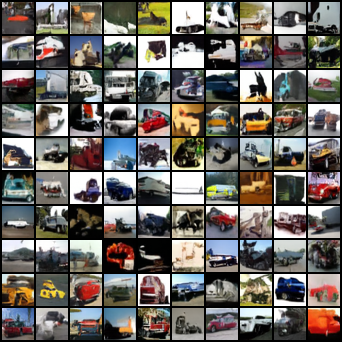}
    }
    \caption{HybViT generated class-conditional samples of \textbf{Truck}}
    \label{figure:hybvit_app_class_9}
\end{figure*}

Following the setting of JEM~\cite{jem}, we conduct a qualitative analysis of samples on CIFAR10. We define an energy function of $\bs{x}$ as $p_{\bs{\theta}}(\bs{x}) \varpropto E(\bs{x}) = \log\! \sum_{y}\!e^{f_{\bs{\theta}}(\bs{x})\left[y\right]}\!=\!\text{LSE}( f_{\bs{\theta}}(\bs{x}))$, the negative of the energy function in~\cite{liu2020energy_ood,jem}. We use a CIFAR10-trained HybViT model to generate 10,000 images from scratch, then feed them back into the HybViT model to compute $E(\bs{x})$ and $p(y|\bs{x})$. We show the examples and distribution by class in Fig~\ref{figure:categorial_topk_bottomk} and Fig~\ref{fig:px_category}.  
We can observe that the worst examples of Plane can be completely blank.
Additional HybViT generated class-conditional (best and worst) samples of CIFAR10 are provided in Figures~\ref{figure:hybvit_app_class_0}-\ref{figure:hybvit_app_class_9}.

\begin{table}[H]
\vspace{-5pt}
\caption{Ablation Study of Data Augmentation on CIFAR10.}
\vspace{-10pt}
\label{table:augmentation}
\begin{center}
\begin{threeparttable}
\begin{tabular}{lcccc}
\toprule
 Model & Aug         & Acc \% $\uparrow$ & IS $\uparrow$ & FID $\downarrow$ \\
 \midrule
\multirow{2}{*}{ViT}   & Strong      & 96.5 &  -    & -    \\
       &  Weak       & 87.1 &  -    & -    \\
\multirow{2}{*}{HybViT} & Strong      & 95.9 & 7.68  & 26.4 \\
       & Weak        & 84.6 & 7.85  & 24.9 \\
\bottomrule
\end{tabular}
\end{threeparttable}
\vspace{-5pt}
\end{center}
\vspace{-10pt}
\end{table}

\subsection{Data Augmentation}

We study the effect of data augmentation. ViT is known to require a too large amount of training data and/or repeated strong data augmentations to obtain acceptable visual representation.  Table~\ref{table:augmentation} compares the performance between strong augmented data and conventional Inception-style pre-processed(namely weak augmentation) data~\cite{szegedy2016rethinking}. We can conclude that the strong data augmentation is really essential for high classification performance and the effect on generation is negative but tiny. 
Note that the data augmentation is only used for classification, and for DDPM, we don't apply any data augmentation. 

\subsection{Out-of-Distribution Detection}

Another useful OOD score function is the maximum probability from a classifier's predictive distribution: $s_{\bs{\theta}}(\bs{x}) = \max_y p_{\bs{\theta}}(y|\bs{x})$. The results can be found in Table~\ref{table:CIFAR10_ood_app} (bottom row). 

\begin{table*}[ht!]
\vspace{-5pt}
\caption{OOD detection results. Models are trained on CIFAR10. Values are AUROC.}
\label{table:CIFAR10_ood_app}
\begin{center}
\begin{threeparttable}
\begin{tabular}{c|c|cccc}
\toprule
$s_{\bs{\theta}}(\bs{x})$  & Model   & SVHN & CIFAR10 Interp & CIFAR100 & CelebA \\
\midrule
\multirow{8}{*}{$\log p_{\bs{\theta}}(\bs{x})$} 
  & WideResNet~\cite{liu2020energy_ood} & .91 & -   & \bf{.87} & .78 \\
  & IGEBM~\cite{du2019implicit}         & .63 & .70 & .50 & .70 \\
  & JEM~\cite{jem}                      & .67 & .65 & .67 & .75 \\
  & JEM++~\cite{jempp}                  & .85 & .57 & .68 & .80 \\
  & VERA~\cite{nomcmc}                  & .83 & .86 & .73 & .33 \\
  & ImCD~\cite{improvedCD}              & .91 & .65 & .83 & - \\
  &    ViT                              & \bf{.93} & \bf{.93} & .82 & \bf{.81} \\
  & HybViT                              & .93 & .92 & .84 & .76\\
\midrule
\multirow{5}{*}{$\max_y p_{\bs{\theta}}(y|\bs{x})$} 
    & WideResNet                  & .93 & .77 & .85 & .62 \\
    & IGEBM~\cite{du2019implicit} & .43 & .69 & .54 & .69 \\
    & JEM~\cite{jem}              & .89 & .75 & .87 & .79 \\
    & JEM++~\cite{jempp}          & \bf{.94} & .77 & \bf{.88} & \bf{.90} \\
    & ViT                      & .91 & \bf{.95} & .82 & .74 \\
    & HybViT                      & .91 & .94 & .85 & .67 \\
\bottomrule
\end{tabular}
\end{threeparttable}
\end{center}
\vspace{-5pt}
\end{table*}

\subsection{Robustness}

Given ViT models trained with different data augmentations, we can investigate their robustness since weak data augmentations are commonly used in CNNs. Table~\ref{table:cifar10_pgd_inf_app_vit} shows an interesting phenomena that HybViT with weak data augmentation is much robust than other models, especially under $L_2$ attack. We suppose it's because the noising process feeds huge amount of noisy samples to HybViT, then HybViT learns from the noisy data implicitly to improve the flatness and robustness.

\begin{table*}[ht!]
\caption{Classification accuracies when models are under $L_\infty$ and $L_2$ PGD attacks with different $\epsilon$'s. All models are trained on CIFAR10.}\label{table:cifar10_pgd_inf_app_vit}
\begin{center}
\begin{threeparttable}
\begin{tabular}{l|c|cccccccc}
\toprule
Model       & Clean (\%)  & $L_\infty$ $\epsilon=1/255$ & $2$  & $4$   & $8$  & $12$  & $16$ & $22$ & $30$ \\
\midrule
ViT         & 96.5  &  70.8  &  46.7  & 21.7  & 7.0  & 1.4 & 0.1 & 0  & 0 \\
- Weak Aug  & 87.1  &  67.3  &  41.8  & 14.8  & 1.4  & 0.1 & 0   & 0  & 0 \\
HybViT      & 95.9  &  70.4  &  48.0  & 21.9  & 5.5  & 1.3 & 0.3 & 0  & 0 \\
- Weak Aug  & 84.6  &  71.3  &  55.6  & 30.3  & 6.7  & 0.6 & 0.1 & 0  & 0 \\
\midrule
Model       & Clean (\%)  & $L_2$ $\epsilon=50/255$ & $100$  & $150$   & $200$  & $250$  & $300$ & $350$ & $400$ \\
\midrule
ViT         & 96.5  &  52.3  &   9.2  & 1.1   & 0.3  & 0.1 & 0.1 & 0   & 0   \\
- Weak Aug  & 87.1  &  53.9  &  21.4  & 5.5   & 1.0  & 0.1 & 0   & 0   & 0   \\
HybViT      & 95.9  &  58.7  &  16.3  & 3.4   & 1.0  & 0.2 & 0.1 & 0.1 & 0   \\
- Weak Aug  & 84.6  &  65.8  &  42.3  & 25.7  & 13.2 & 6.4 & 3.4 & 1.5 & 0.7 \\
\bottomrule
\end{tabular}
\end{threeparttable}
\end{center}
\end{table*}

\subsection{Calibration}

\begin{figure}[ht!]
    \centering
    \subfigure[WRN (w/o BN)]{
        \includegraphics[width=0.3\columnwidth]{figures/calibration_wrn.png}
        \label{figure:base_cali_app}
    }
    \subfigure[JEM]{
        \includegraphics[width=0.3\columnwidth]{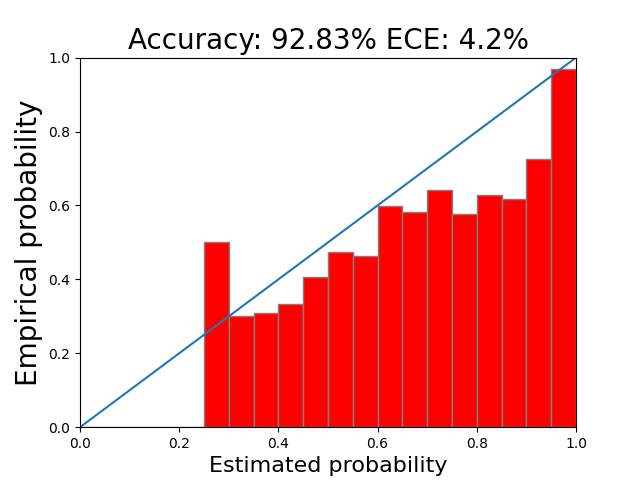}
        \label{figure:jem_cali_app}
    }
    \subfigure[ViT]{
        \includegraphics[width=0.3\columnwidth]{figures/calibration_vit.png}
        \label{figure:vit_cali_app}
    }
    \subfigure[HybViT]{
        \includegraphics[width=0.3\columnwidth]{figures/calibration_hybvit.png}
        \label{figure:hybvit_cali_app}
    }
    \subfigure[ViT w/ Weak Aug]{
        \includegraphics[width=0.3\columnwidth]{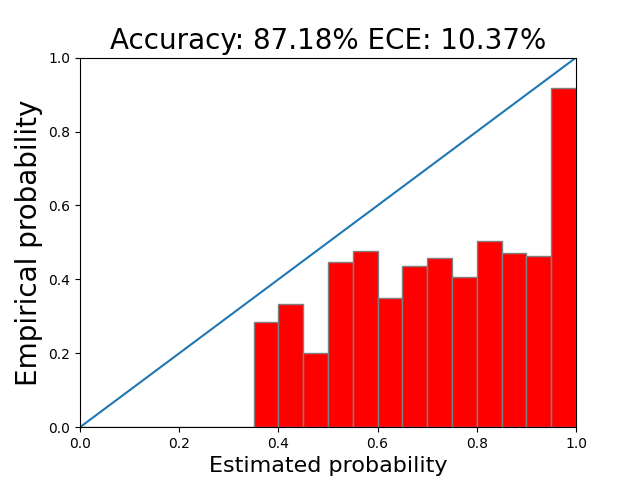}
        \label{figure:vit_weak_cali_app}
    }
    \subfigure[HybViT w/ Weak Aug]{
        \includegraphics[width=0.3\columnwidth]{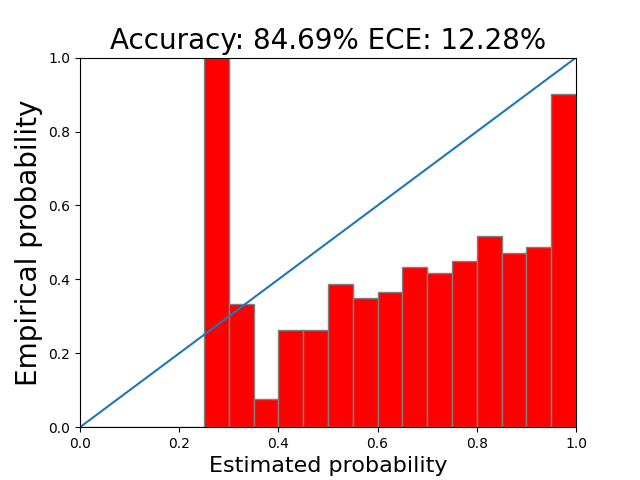}
        \label{figure:hybvit_weak_cali_app}
    }
    \vspace{-5pt}
    \caption{Calibration results on CIFAR10. The smaller ECE is, the better.}
    \label{figure:CIFAR10_cali_app_hybvit}
\end{figure}

Figures in~\ref{figure:CIFAR10_cali_app_hybvit} provide a comparison of ViT and HybViT with the baselines WRN and JEM, and also corresponding ViTs trained without strong data augmentations. It can be observed that strong data augmentations can better calibrate the predictions of ViT and HybViT, but further make them under-confident.

\subsection{Training Speed}

We further report the empirical training speeds of our models and baseline methods for ImageNet 32x32. Our methods are memory efficient since it only requires a single GPU, and much faster.

\begin{table}[!ht]
\caption{Run-time comparison on ImageNet 32x32. All experiments are performed on a single GPU for 100 epochs.}
\vspace{-10pt}

\label{table:speed_img32}
\begin{center}

\begin{threeparttable}
\begin{tabular}{lcc}
\toprule
Model              & NoP(M)  &  Runtime   \\
\midrule
ViT                & 11.6 &  3 days  \\
GenViT             & 11.6 &  2 days  \\
HybViT             & 11.6 &  5 days  \\
\midrule
IGEBM              &  \multicolumn{2}{c}{32 GPUs for 5 days}              \\
KL-EBM             &  \multicolumn{2}{c}{8 GPUs for 3 days}        \\
\bottomrule
\end{tabular}
\end{threeparttable}
\vspace{-10pt}
\end{center}
\end{table}

\begin{figure}[ht]

    \centering
    \caption{Additional generated images on benchmark datasets}

    \subfigure[CIFAR10]{
        \includegraphics[width=0.3\columnwidth]{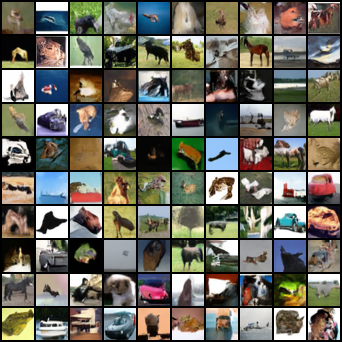} \label{figure:cifar10_app}
    }
    \subfigure[CIFAR100]{
        \includegraphics[width=0.3\columnwidth]{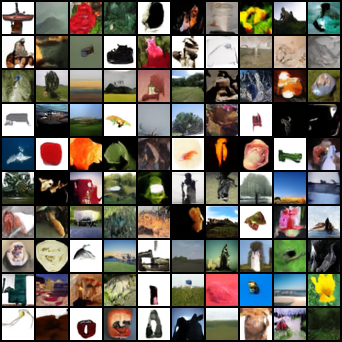} \label{figure:cifar100_app}
    }
    \subfigure[ImageNet 32x32]{
        \includegraphics[width=0.3\columnwidth]{figures/extra_generated/cifar100_hybvit.png} \label{figure:img32_app}
    }
    \subfigure[TinyImagenet]{
        \includegraphics[width=0.3\columnwidth]{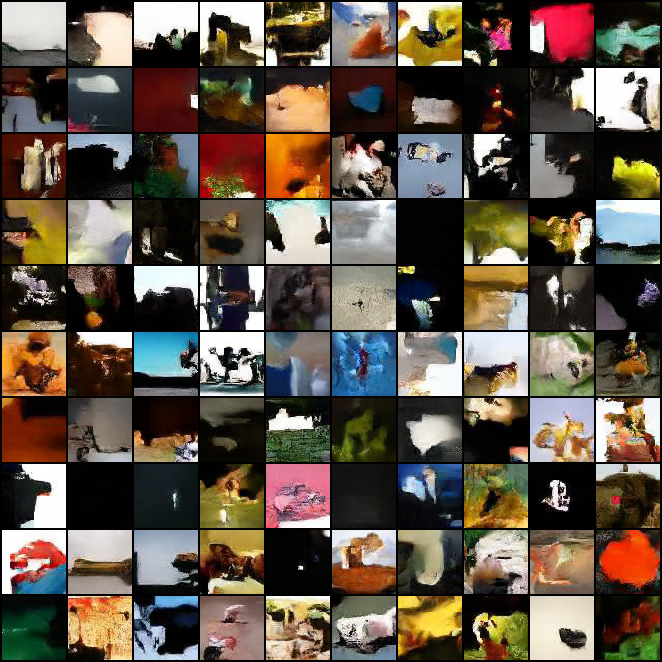} \label{figure:tinyimg_app}
    }
    \subfigure[STL10]{
        \includegraphics[width=0.3\columnwidth]{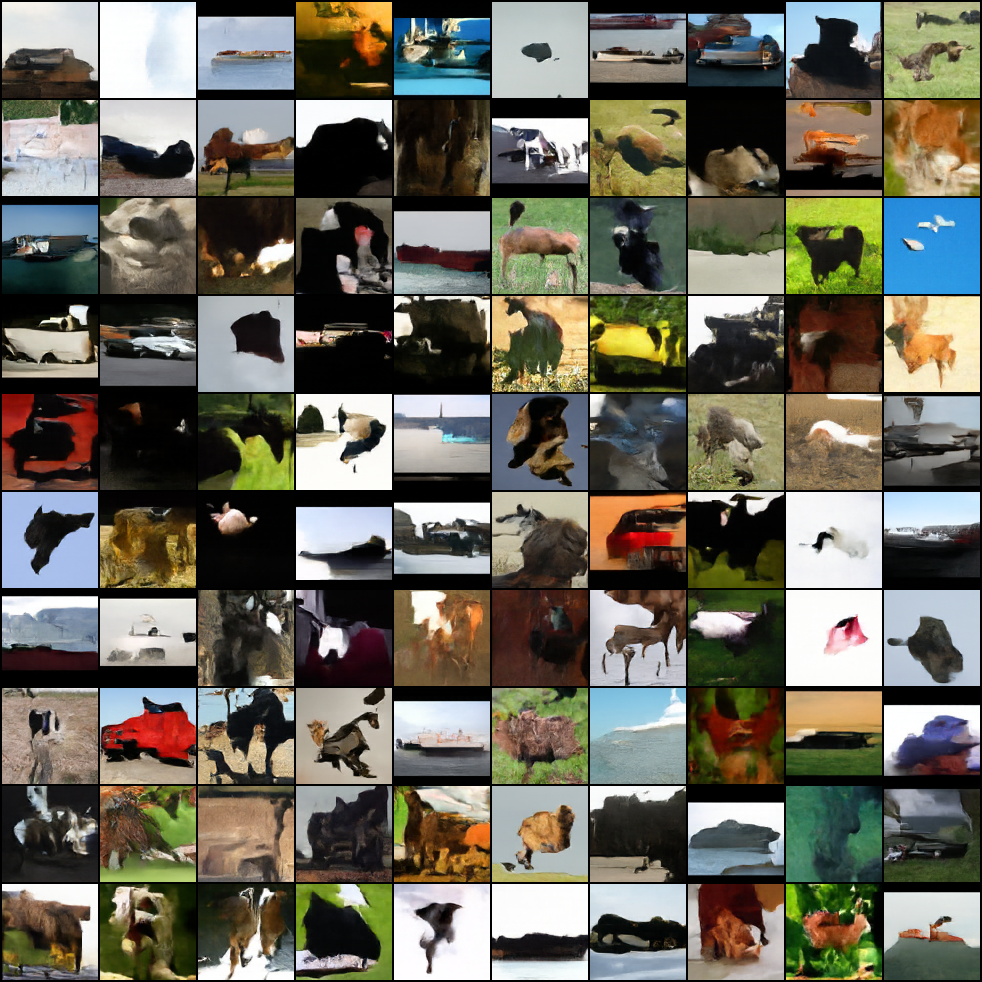} \label{figure:stl10_app}
    }
    \subfigure[CelebA 128]{
        \includegraphics[width=0.3\columnwidth]{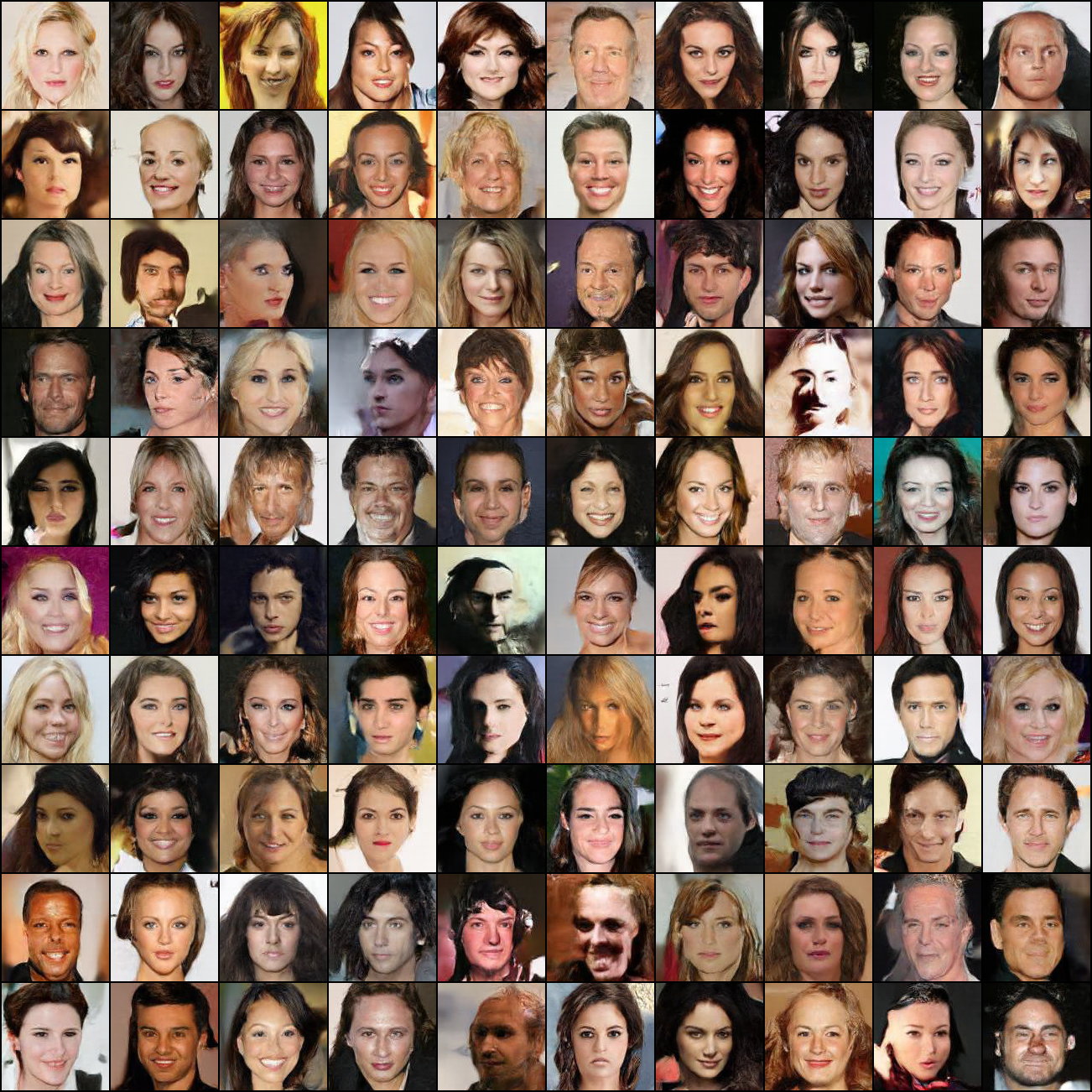} \label{figure:celeba128_app}
    }
    \label{figure:generated_images_app}

\end{figure}

\section{Additional Generated Samples}

\begin{figure}[ht]
    \centering
    \subfigure[ImageNet 128x128]{
        \includegraphics[width=0.35\columnwidth]{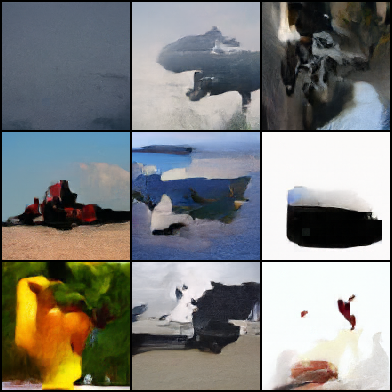} \label{figure:img128_app}
    }
    \subfigure[ImageNet 224x224]{
        \includegraphics[width=0.35\columnwidth]{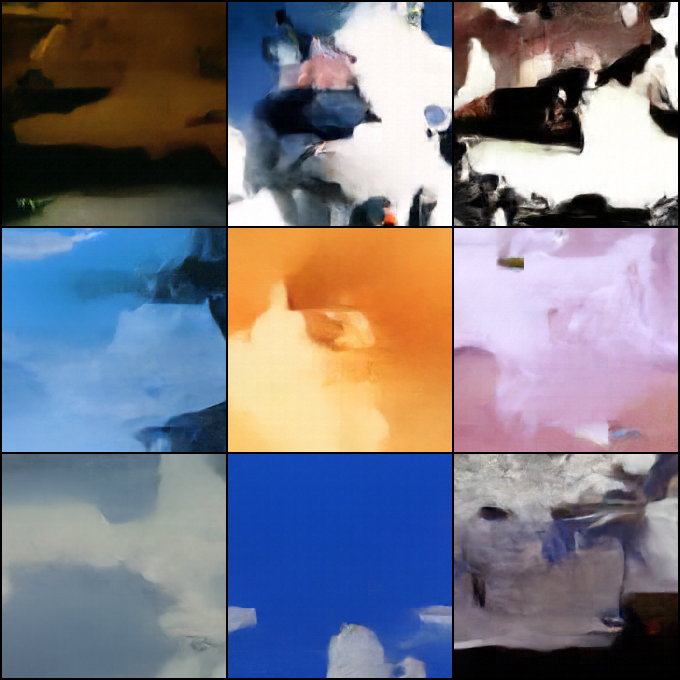} \label{figure:img224_app}
    }
    \caption{Generated Images}
    \label{figure:generated_imagenet_app}
\end{figure}

Additional generated samples of CIFAR10, CIFAR100, ImageNet 32x32, TinyImageNet, STL10, and CelebA 128 are provided in Figure~\ref{figure:generated_images_app}.  We further provide some generated images for ImageNet 128x128 and vanilla ImageNet 224x224 are shown in \ref{figure:generated_imagenet_app}. , The patch size are set as 8 and 14 for ImageNet 128 and 224 respectively. Similar to previous discussion about patch size, we find the generation quality is very low. Due to limited computation resource and low generation quality, we only show a preliminary generative results on ImageNet-128 and vanilla ImageNet 224x224.

\end{document}